\documentclass{article}
\usepackage{iclr2023,times}

\usepackage{amsmath,amsfonts,bm}

\def\eqref#1{equation~\ref{#1}}

\def\1{\bm{1}}

\DeclareMathAlphabet{\mathsfit}{\encodingdefault}{\sfdefault}{m}{sl}
\SetMathAlphabet{\mathsfit}{bold}{\encodingdefault}{\sfdefault}{bx}{n}

\usepackage{url}
\usepackage{rotating}

\usepackage{subfigure}
\usepackage{booktabs}
\usepackage{amsmath,bm}
\usepackage{bbm}
\usepackage{cite}
\usepackage{enumerate}
\usepackage{algorithm}
\usepackage{algorithmic}
\usepackage{colortbl}
\usepackage{psfrag}
\usepackage{adjustbox}
\usepackage{wrapfig}
\usepackage{xspace} 
\usepackage{amssymb}
\usepackage{multirow}
\usepackage{afterpage}
\usepackage{xcolor}
\usepackage{float}

\usepackage{enumitem}

\newenvironment{Itemize}{
    \begin{itemize}[leftmargin=*]
    \setlength{\itemsep}{0pt}
    \setlength{\topsep}{0pt}
    \setlength{\partopsep}{0pt}
    \setlength{\parskip}{0pt}}
{\end{itemize}}
\usepackage[most]{tcolorbox}
\tcbset{
    frame code={}
    center title,
    left=0pt,
    right=0pt,
    top=0pt,
    bottom=0pt,
    colframe=white,
    width=\dimexpr\textwidth\relax,
    enlarge left by=0mm,
    boxsep=5pt,
    arc=0pt,outer arc=0pt,
}

\usepackage{xcolor}
\colorlet{darkgreen}{green!65!black}
\colorlet{darkblue}{blue!75!black}
\colorlet{darkred}{red!80!black}
\definecolor{lightblue}{HTML}{0071bc}
\definecolor{lightgreen}{HTML}{39b54a}
\definecolor{manyshot}{HTML}{6969ff}
\definecolor{medshot}{HTML}{f7c600}
\definecolor{fewshot}{HTML}{ff6969}
\definecolor{mypurple}{HTML}{412F8A}
\definecolor{myorange}{HTML}{fc8e62}

\definecolor{deemph}{gray}{0.55}
\newcommand{\gc}[1]{\textcolor{deemph}{#1}}

\usepackage[pagebackref=false, breaklinks=true, colorlinks,
            citecolor=citecolor, linkcolor=linkcolor, bookmarks=false]{hyperref}
\definecolor{citecolor}{HTML}{0071BC}
\definecolor{linkcolor}{HTML}{ED1C24}

\global\long\def\real{\mathbb{R}}

\renewcommand{\hat}{\widehat}

\newcommand{\simper}{\texttt{SimPer}\xspace}

\renewcommand{\paragraph}[1]{\vspace{1.25mm}\noindent\textbf{#1}}
\newcommand{\grayrow}{\rowcolor[gray]{.9}}
\definecolor{baselinecolor}{gray}{.95}
\newcommand{\graycell}[1]{\cellcolor{baselinecolor}{#1}}

\title{SimPer: Simple Self-Supervised Learning of \\ Periodic Targets}

\author{Yuzhe Yang$^{1}$, Xin Liu$^{2}$, Jiang Wu$^{3}$, Silviu Borac$^{3}$, Dina Katabi$^{1}$, Ming-Zher Poh$^{3}$, Daniel McDuff$^{2,3}$ \\[2.5pt]
$^{1}$MIT CSAIL \quad $^{2}$University of Washington \quad $^{3}$Google
}

\iclrfinalcopy  

\begin{document}

\maketitle

\vspace{-0.1cm}
\begin{abstract}
From human physiology to environmental evolution, important processes in nature often exhibit meaningful and strong \emph{periodic} or \emph{quasi-periodic} changes.
Due to their inherent label scarcity, learning useful representations for periodic tasks with limited or no supervision is of great benefit.
Yet, existing self-supervised learning (SSL) methods overlook the intrinsic periodicity in data, and fail to learn representations that capture periodic or frequency attributes.
In this paper, we present \simper, a simple contrastive SSL regime for learning periodic information in data.
To exploit the periodic inductive bias, \simper introduces customized augmentations, feature similarity measures, and a generalized contrastive loss for learning efficient and robust periodic representations.
Extensive experiments on common real-world tasks in human behavior analysis, environmental sensing, and healthcare domains verify the superior performance of \simper compared to state-of-the-art SSL methods, highlighting its intriguing properties including better data efficiency, robustness to spurious correlations, and generalization to distribution shifts.
Code and data are available at: {\url{https://github.com/YyzHarry/SimPer}}.
\end{abstract}

\begin{figure}[h]
\begin{center}
\includegraphics[width=\linewidth]{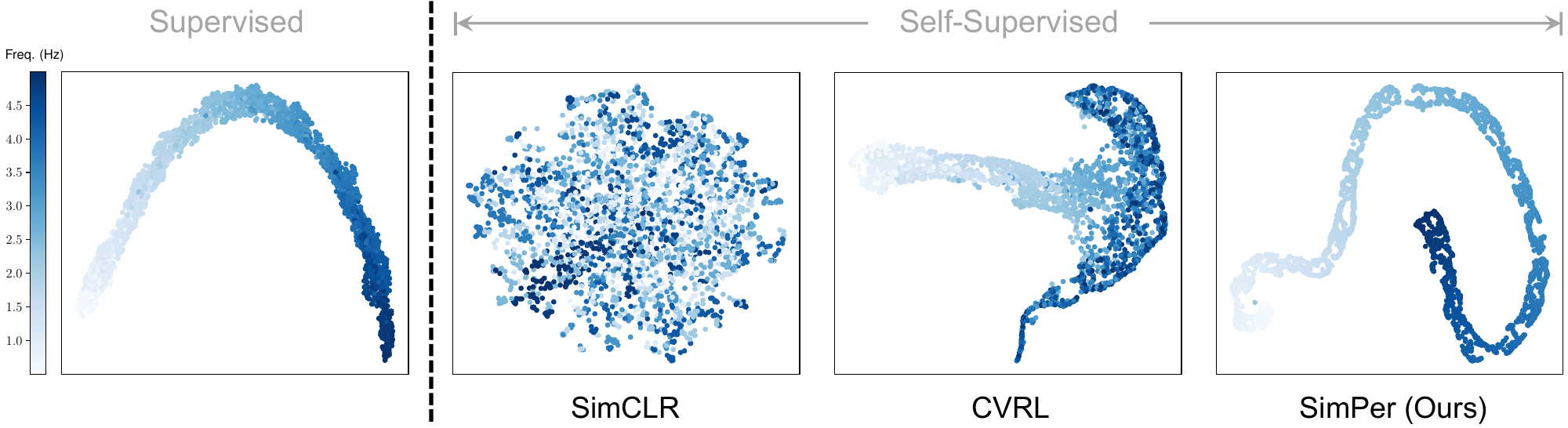}
\end{center}
\vspace{-0.4cm}
\caption{\small
Learned representations of different methods on a periodic learning dataset, RotatingDigits~(details in Section~\ref{sec:experiment}). Existing self-supervised learning schemes fail to capture the underlying periodic or frequency information in data. In contrast, \simper learns robust periodic representations with high frequency resolution.}
\label{fig:teaser-intro}
\vspace{-0.1cm}
\end{figure}

\vspace{-0.1cm}
\section{Introduction}
\vspace{-0.1cm}
\label{sec:intro}
Practical and important applications of machine learning in the real world, from monitoring the earth from space using satellite imagery \citep{espeholt2021skillful} to detecting physiological vital signs in a human being \citep{luo2019smartphone}, often involve recovering \emph{\textbf{periodic}} changes.
In the \textbf{health} domain, learning from video measurement has shown to extract (quasi-)periodic vital signs including atrial fibrillation \citep{yan2018contact}, sleep apnea episodes \citep{amelard2018non} and blood pressure \citep{luo2019smartphone}.
In the \textbf{environmental remote sensing} domain, periodic learning is often needed to enable nowcasting of environmental changes such as precipitation patterns or land surface temperature \citep{sonderby2020metnet}.
In the \textbf{human behavior analysis} domain, recovering the frequency of changes or the underlying temporal morphology in human motions (e.g., gait or hand motions) is crucial for those rehabilitating from surgery \citep{gu2019home}, or for detecting the onset or progression of neurological conditions such as Parkinson's disease \citep{doi:10.1126/scitranslmed.adc9669,yang2022artificial}.

While learning periodic targets is important, labeling such data is typically challenging and resource intensive. For example, if designing a method to measure heart rate, collecting videos with highly synchronized gold-standard signals from a medical sensor is time consuming, labor intensive, and requires storing privacy sensitive bio-metric data.
Fortunately, given the large amount of unlabeled data, \emph{self-supervised learning} that captures the underlying periodicity in data would be promising.

Yet, despite the great success of self-supervised learning (SSL) schemes on solving discrete classification or segmentation tasks, such as image classification \citep{chen2020simple,he2020momentum}, object detection \citep{xiao2021region}, action recognition \citep{qian2021spatiotemporal}, or semantic labeling \citep{hu2021region}, less attention has been paid to designing algorithms that capture periodic or quasi-periodic temporal dynamics from data. 
Interestingly, we highlight that existing SSL methods inevitably overlook the intrinsic periodicity in data: Fig.~\ref{fig:teaser-intro} shows the UMAP \citep{mcinnes2018umap} visualization of learned representations on RotatingDigits, a toy periodic learning dataset that aims to recover the underlying rotation frequency of different digits (details in Section~\ref{sec:experiment}).
As the figure shows, state-of-the-art (SOTA) SSL schemes fail to capture the underlying periodic or frequency information in the data.
Such observations persist across tasks and domains as we show later in Section~\ref{sec:experiment}.



To fill the gap, we present \simper, a simple self-supervised regime for learning periodic information in data.
Specifically, to leverage the temporal properties of periodic targets, \simper first introduces a \emph{temporal self-contrastive learning} framework, where positive and negative samples are obtained through \emph{periodicity-invariant} and \emph{periodicity-variant} augmentations from the \textbf{same} input instance.
Further, we identify the problem of using conventional feature similarity measures (e.g., $\cos(\cdot)$) for periodic representation, and propose \emph{periodic feature similarity} to explicitly define how to measure similarity in the context of periodic learning.
Finally, to harness the intrinsic \emph{continuity} of augmented samples in the frequency domain, we design a \emph{generalized contrastive loss} that extends the classic InfoNCE loss to a soft regression variant that enables contrasting over continuous labels (frequency).

To support practical evaluation of SSL of periodic targets, we benchmark \simper against SOTA SSL schemes on six diverse periodic learning datasets for common real-world tasks in human behavior analysis, environmental remote sensing, and healthcare. Rigorous experiments verify the robustness and efficiency of \simper on learning periodic information in data.
Our contributions are as follows:
\vspace{-15pt}
\begin{Itemize}
    \item We identify the limitation of current SSL methods on periodic learning tasks, and uncover intrinsic properties of learning periodic dynamics with self-supervision over other mainstream tasks.
    \vspace{1pt}
    \item We design \simper, a simple \& effective SSL framework that learns periodic information in data.
    \vspace{1pt}
    \item We conduct extensive experiments on six diverse periodic learning datasets in different domains: human behavior analysis, environmental sensing, and healthcare. Rigorous evaluations verify the superior performance of \simper against SOTA SSL schemes.
    \vspace{1pt}
    \item Further analyses reveal intriguing properties of \simper on its data efficiency, robustness to spurious correlations \& reduced training data, and generalization to unseen targets.
\end{Itemize}

\vspace{-0.3cm}
\section{Related Work}
\vspace{-0.1cm}
\label{sec:related-work}
\textbf{Periodic Tasks in Machine Learning.}
Learning or recovering periodic signals from high dimensional data is prevailing in real-world applications.
Examples of periodic learning include recovering and magnifying physiological signals (e.g., heart rate or breathing) \citep{wu2012eulerian}, predicting weather and environmental changes (e.g., nowcasting of precipitation or land surface temperatures) \citep{sonderby2020metnet,espeholt2021skillful}, counting motions that are repetitious (e.g., exercises or therapies) \citep{dwibedi2020counting,ali2020spatio}, and analyzing human behavior (e.g., gait) \citep{doi:10.1126/scitranslmed.adc9669}.
To date, much prior work has focused on designing customized neural architectures \citep{liu2020multi,dwibedi2020counting}, loss functions \citep{starke2022deepphase}, and leveraging relevant learning paradigms including transfer learning \citep{lu2018class} and meta-learning \citep{liu2021metaphys} for periodic learning in a \emph{supervised} manner, with high-quality labels available.
In contrast to these past work, we aim to learn robust \& efficient periodic representations in a \emph{self-supervised} manner.

\textbf{Self-Supervised Learning.}
Learning with self-supervision has recently attracted increasing interests, where early approaches mainly rely on pretext tasks, including exemplar classification \citep{dosovitskiy2014discriminative}, solving jigsaw puzzles \citep{noroozi2016unsupervised}, object counting \citep{noroozi2017representation}, clustering \citep{caron2018deep}, and predicting image rotations \citep{gidaris2018unsupervised}.
More recently, a line of work based on contrastive losses \citep{oord2018representation,tian2019contrastive,chen2020simple,he2020momentum} shows great success in self-supervised representations, where similar embeddings are learned for different views of the same training example (\emph{positives}), and dissimilar embeddings for different training examples (\emph{negatives}).
Successful extensions have been made to temporal learning domains including video understanding \citep{jenni2020video} or action classification \citep{qian2021spatiotemporal}.
However, current SSL methods have limitations in learning periodic information, as the periodic inductive bias is often overlooked in method design. Our work extends existing SSL frameworks to periodic tasks, and introduces new techniques suitable for learning periodic targets.

\section{The SimPer Framework}
\vspace{-0.2cm}
\label{sec:method}
\begin{figure}[!t]
\begin{center}
\includegraphics[width=\linewidth]{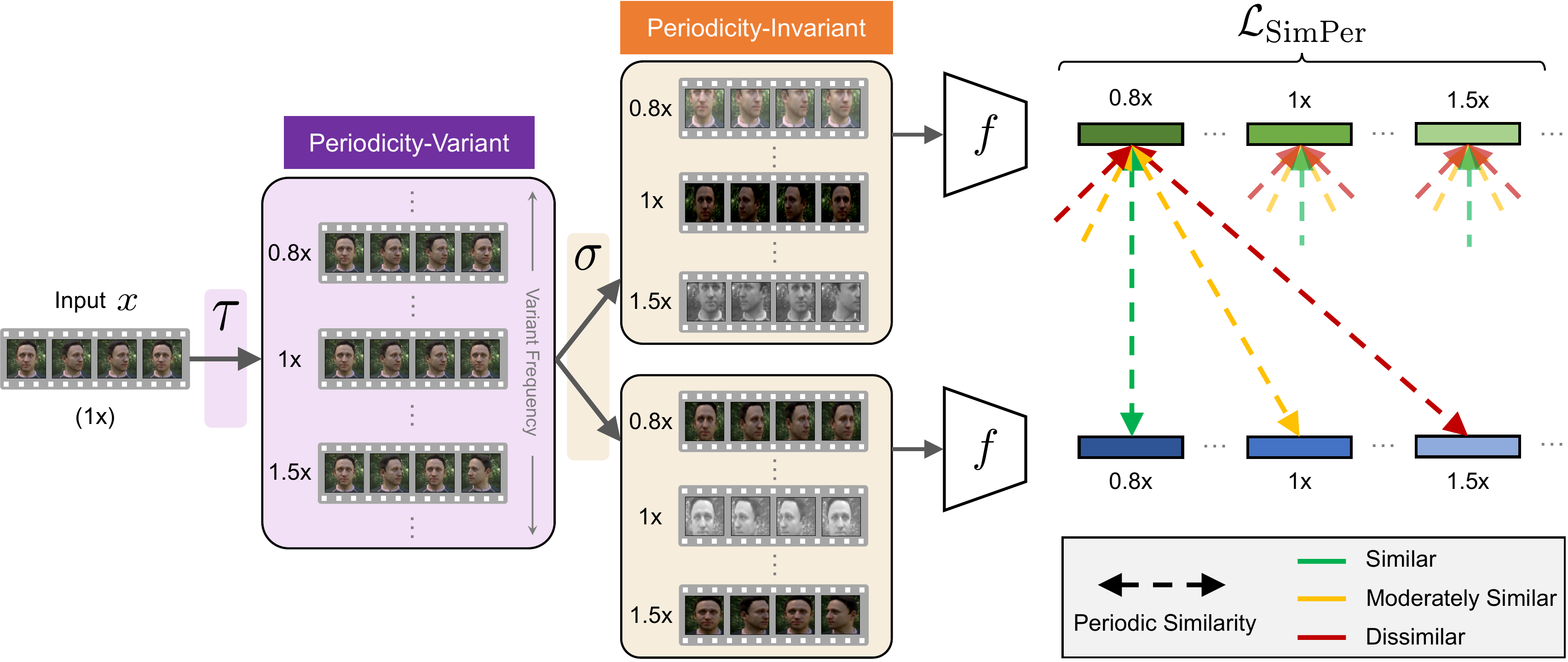}
\end{center}
\vspace{-0.4cm}
\caption{\small
\textbf{An overview of the \simper framework.} Input sequence is first passed through periodicity-variant transformations $\tau(\cdot)$ to create a series of speed (frequency) changed samples, where each augmented sample exhibits different underlying periodic signals due to the altered frequency, and can be treated as \emph{negative} examples for each other. The augmented series are then passed through two sets of periodicity-invariant transformations $\sigma(\cdot)$ to create different invariant views (\emph{positives}). All samples are then encoded in the feature space through a shared encoder $f(\cdot)$. The \simper loss is calculated by contrasting over continuous speed (frequency) labels of different feature vectors, using customized periodic feature similarity measures.}
\label{fig:simper-overview}
\vspace{-0.3cm}
\end{figure}

When learning from periodic data in a self-supervised manner, a fundamental question arises:
\vspace{-0.25cm}
\begin{center}
    \emph{How do we design a self-supervised task such that \textbf{periodic} inductive biases are exploited?}
\end{center}
\vspace{-0.25cm}

We note that periodic learning exhibits characteristics that are distinct from prevailing learning tasks.
\emph{First}, while most efforts on exploring invariances engineer transformations in the spatial (e.g., image recognition) or temporal (e.g., video classification) domains, dynamics in the \textbf{{frequency}} domain are essential in periodic tasks, which has implications for how we design (in)variances.
\emph{Second}, unlike conventional SSL where a cosine distance is typically used for measuring feature similarity, repre-sentations learned for repetitious targets inherently possess periodicity that is insensitive to certain shifts (e.g., shifts in feature index), which warrants new machinery for measuring \emph{periodic} similarity.
\emph{Third}, labels of periodic data have a natural ordinality and continuity in the frequency domain, which inspires the need for strategies beyond instance discrimination, that contrast over \emph{continuous} targets.


We present \simper (\underline{Sim}ple SSL of \underline{Per}iodic Targets), a unified SSL framework that addresses each of the above limitations.
Specifically, \simper first introduces a \emph{\textbf{temporal self-contrastive learning}} scheme, where we design \emph{periodicity-invariant} and \emph{periodicity-variant} augmentations for the \emph{same} input instance to create its effective \emph{positive} and \emph{negative} views in the context of periodic learning (Section \ref{subsec:temporal-self-con-framework}).
Next, \simper presents \emph{\textbf{periodic feature similarity}} to explicitly define how one should measure the feature similarity when the learned representations inherently possess periodic information (Section \ref{subsec:period-feat-sim}).
Finally, in order to exploit the continuous nature of augmented samples in the frequency domain, we propose a \emph{\textbf{generalized contrastive loss}} that extends the classic InfoNCE loss \citep{oord2018representation} from \emph{discrete} instance discrimination to \emph{continuous} contrast over frequencies, which takes into account the meaningful distance between continuous labels (Section \ref{subsec:generalized-con-loss}).


\vspace{-0.1cm}
\subsection{Temporal Self-Contrastive Learning Framework}
\vspace{-0.1cm}
\label{subsec:temporal-self-con-framework}

\textbf{Problem Setup.}
Let $\mathcal{D} = \{ ( \mathbf{x}_i)\}_{i=1}^{N}$ be the unlabeled training set, where $\mathbf{x}_i\in\real^{D}$ denotes the input sequence.
We denote as $\mathbf{z} = f(\mathbf{x};\theta)$ the representation of $\mathbf{x}$, where $f(\cdot;\theta)$ is parameterized by a deep neural network with parameter $\theta$.
To preserve the full temporal dynamics and information, $\mathbf{z}$ typically extracts \emph{frame-wise} feature of $\mathbf{x}$, i.e., $\mathbf{z}$ has the same length as input $\mathbf{x}$.

As motivated, \emph{\textbf{frequency}} information is most essential when learning from periodic data. Precisely, augmentations that change the underlying frequency effectively alter the identity of the data (periodicity), and vice versa. This simple insight has implications for how we design proper (in)variances.

\textbf{Periodicity-Variant Augmentations.}
We construct negative views of data through transformations in the frequency domain.
Specifically, given input sequence $\mathbf{x}$, we define periodicity-variant augmentations $\tau \in \mathcal{T}$, where $\mathcal{T}$ represents the set of transformations that change $\mathbf{x}$ with an \emph{arbitrary} speed that is feasible under the Nyquist sampling theorem.
As Fig.~\ref{fig:simper-overview} shows, \simper augments $\mathbf{x}$ by $M$ times, obtaining a series of \emph{speed} (\emph{frequency}) changed samples $\{\tau_1(\mathbf{x}),\tau_2(\mathbf{x})\},\dots,\tau_M(\mathbf{x})\}$, whose relative speeds satisfy $s_1 < s_2 < ...<s_M, s_i\propto\text{freq}(\tau_i(\mathbf{x}))$.
Such augmentation effectively changes the underlying periodic targets with shifted frequencies, thus creating different \emph{negative} views.
Therefore, although the original target frequency is unknown, we effectively devise \emph{pseudo speed (frequency) labels} for unlabeled $\mathbf{x}$.
In practice, we limit the speed change range to be within $[s_{\text{min}}, s_{\text{max}}]$, ensuring the augmented sequence is longer than a fixed length in the time dimension.

\textbf{Periodicity-Invariant Augmentations.}
We further define periodicity-invariant augmentation $\sigma \in \mathcal{S}$, where $\mathcal{S}$ denotes the set of transformations that do not change the \emph{identity} of the original input.
When the set is finite, i.e., $\mathcal{S} = \{\sigma_1, \dots, \sigma_k\}$, we have $\text{freq}(\sigma_i(\mathbf{x})) = \text{freq}(\sigma_j(\mathbf{x})), \forall i,j\in[k]$.
Such augmentations can be used to learn invariances in the data from the perspective of periodicity, creating different \emph{positive} views.
Practically, we leverage spatial (e.g., crop \& resize) and temporal (e.g., reverse, delay) augmentations to create different views of the same instance (see Fig.~\ref{fig:simper-overview}).

\setlength\intextsep{-4pt}
\begin{wraptable}[7]{r}{0.4\textwidth}
\setlength{\tabcolsep}{4pt}
\caption{\small Differences of view constructions.}
\vspace{-10pt}
\label{table:pos-neg-differences}
\small
\begin{center}
\adjustbox{max width=0.4\textwidth}{
\begin{tabular}{lcc}
\toprule[1.5pt]
\textbf{Algorithm} & \textbf{Positives} & \textbf{Negatives} \\
\midrule
\begin{tabular}[c]{@{}l@{}}Conventional\\ SSL methods\end{tabular} &
\begin{tabular}[c]{@{}c@{}} \textbf{Instance:} \color{darkgreen}Same \\[1.5pt] \textbf{Aug.:} Invariant\end{tabular} &
\graycell{\begin{tabular}[c]{@{}c@{}} \textbf{Instance:} \color{darkred}Different \\[1.5pt] \textbf{Aug.:} Invariant \end{tabular}} \\
\midrule
\simper &
\begin{tabular}[c]{@{}c@{}} \textbf{Instance:} \color{darkgreen}Same \\[1.5pt] \textbf{Aug.:} \color{darkgreen}Period.-Invariant \end{tabular} &
\graycell{\begin{tabular}[c]{@{}c@{}} \textbf{Instance:} \color{darkgreen}Same \\[1.5pt] \textbf{Aug.:} \color{darkred}Period.-Variant \end{tabular}} \\
\bottomrule[1.5pt]
\end{tabular}}
\end{center}
\end{wraptable}

\textbf{Temporal Self-Contrastive Learning.}
Unlike conventional contrastive SSL schemes where augmentations are exploited to produce invariances, i.e., creating different {positive} views of the data, \simper introduces periodicity-variant augmentations to explicitly model what \emph{variances} should be in periodic learning. Concretely, {negative} views are no longer from {other} \emph{different} instances, but directly from the \emph{same} instance itself, realizing a \emph{\textbf{self-contrastive}} scheme.
Table \ref{table:pos-neg-differences} details the differences.

We highlight the benefits of using the self-contrastive framework. First, it provides \emph{arbitrarily large} negative sample sizes, as long as the Nyquist sampling theorem is satisfied. This makes \simper not dependent on the actual training set size, and enables effective contrasting even under limited data scenarios. We show in Section \ref{subsec:exp-data-efficiency} that when drastically reducing the dataset size to only $5\%$ of the total samples, \simper still works equally well, substantially outperforming supervised counterparts.
Second, our method naturally leads to \emph{hard} negative samples, as periodic information is directly being contrasted, while unrelated information (e.g., frame appearance) are maximally preserved across negative samples. This makes \simper robust to spurious correlations in data (Section \ref{subsec:exp-spurious-corr}).

\vspace{-0.2cm}
\subsection{Feature Similarity in the Context of Periodic Learning}
\vspace{-0.15cm}
\label{subsec:period-feat-sim}

We identify that \emph{feature similarity measures} are also different in the context of periodic representations. Consider sampling two short clips $\mathbf{x}_1, \mathbf{x}_2$ from the same input sequence, but with a frame shift $t$. Assume the frequency does not change within the sequence, and its period $T > t$.
Since the underlying information does not change, by definition their features should be close in the embedding space (i.e., high feature similarity).
However, due to the shift in time, when extracting frame-level feature vectors, the indexes of the feature representations (which represent different time stamps) will no longer be aligned. 
In this case, if directly using a cosine similarity as defined in conventional SSL literature, the similarity score would be low, despite the fact that the actual similarity is high.

\begin{figure}[!t]
\begin{center}
\includegraphics[width=\linewidth]{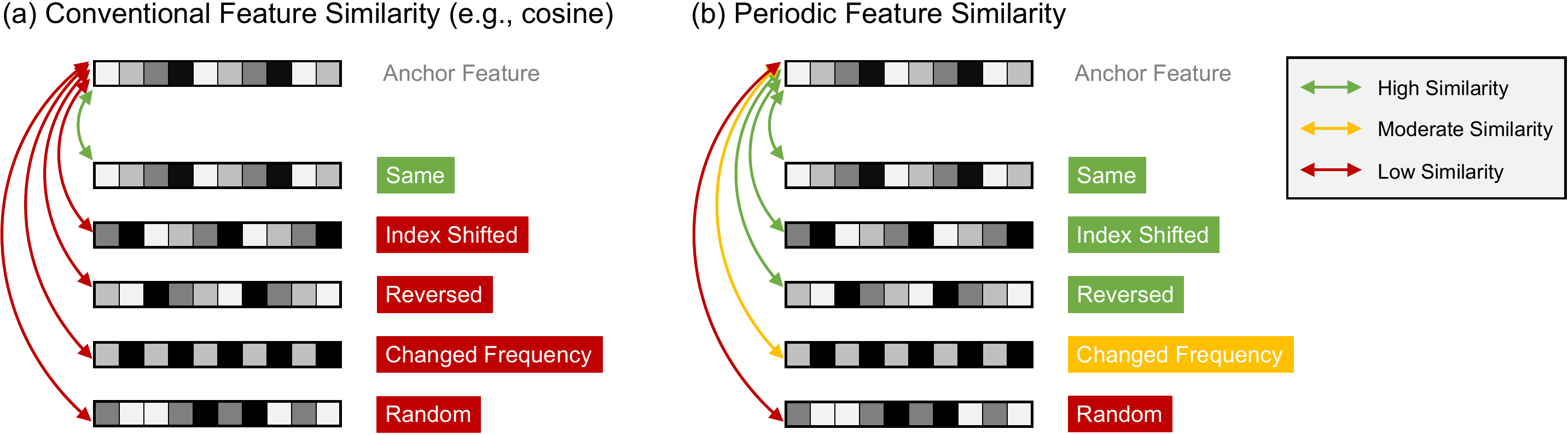}
\end{center}
\vspace{-0.3cm}
\caption{\small
Differences between \textbf{(a)} conventional feature similarity, and \textbf{(b)} \emph{periodic} feature similarity. A proper periodic feature similarity measure should induce high similarity for features with shifted (sometimes reversed) indexes, while capturing a continuous similarity change when the feature frequency varies.
}
\label{fig:periodic-feat-sim}
\vspace{-0.2cm}
\end{figure}

\textbf{Periodic Feature Similarity.}
To overcome this limitation, we propose to use \emph{periodic} feature similarity measures in \simper.
Fig.~\ref{fig:periodic-feat-sim} highlights the properties and differences between conventional feature similarity measures and the desired similarity measure in periodic learning.
Specifically, existing SSL methods adopt similarity measures that emphasize strict ``closeness'' between two feature vectors, and are sensitive to shifted or reversed feature indexes.
In contrast, when aiming for learning periodic features, a proper periodic feature measure should retain high similarity for features with shifted (sometimes reversed) indexes, while also capturing a continuous similarity change when the feature frequency varies, due to the meaningful distance in the frequency domain.

\textbf{Concrete Instantiations.}
We provide two practical instantiations to effectively capture the periodic feature similarity. Note that these instantiations can be easily extended to high-dimensional features (in addition to the time dimension) by averaging across other dimensions.
\vspace{-0.2cm}
\begin{Itemize}
    \item \emph{Maximum cross-correlation (\textbf{MXCorr})} measures the maximum similarity as a function of offsets between signals \citep{welch1974lower}, which can be efficiently computed in the frequency domain.
    \vspace{0.1cm}
    \item \emph{Normalized power spectrum density (\textbf{nPSD})} calculates the distance between the normalized PSD of two feature vectors. The distance can be a cosine or $L_2$ distance (details in Appendix \ref{appendix-subsubsec:similarity-metrics}).
\end{Itemize}
\vspace{-0.1cm}

\vspace{-0.2cm}
\subsection{Generalized Contrastive Loss with Continuous Targets}
\vspace{-0.15cm}
\label{subsec:generalized-con-loss}

Motivated by the fact that the augmented views are \emph{\textbf{continuous}} in frequency, where the \emph{pseudo speed labels} $\{s_i\}_{i=1}^M$ are known through augmentation (i.e., a view at $1.1\times$ is more similar to the original than that at $2\times$), we relax and extend the original InfoNCE contrastive loss \citep{oord2018representation} to a soft variant, where it generalizes from discrete instance discrimination to continuous targets.

\textbf{From Discrete Instance Discrimination to Continuous Contrast.}
The classic formulation of the InfoNCE contrastive loss for each input sample $\mathbf{x}$ is written as
\begingroup\makeatletter\def\f@size{8}\check@mathfonts\def\maketag@@@#1{\hbox{\m@th\large\normalfont#1}}
\begin{equation}
\label{eqn:infonce}
\mathcal{L}_{\text{InfoNCE}}
= - \log \frac{\exp (\operatorname{sim}(\mathbf{z}, \mathbf{\hat{z}}) / \nu)}{\sum_{\mathbf{z}' \in \mathcal{Z} \setminus \{\mathbf{z}\}} \exp (\operatorname{sim}(\mathbf{z}, \mathbf{z}') / \nu)},
\end{equation}
\endgroup
where $\mathbf{\hat{z}}=f(\mathbf{\hat{x}})$ ($\mathbf{\hat{x}}$ is the positive pair of $\mathbf{x}$ obtained through augmentations),
$\mathcal{Z}$ is the set of features in current batch,
$\nu$ is the temperature constant, and $\operatorname{sim}(\cdot)$ is usually instantiated by a dot product.
Such format indicates a \emph{hard} classification task, where target label is $1$ for positive pair and $0$ for all negative pairs.
However, negative pairs in \simper inherently possess a meaningful distance, which is reflected by the similarity of their relative speed (frequency).
To capture this intrinsic continuity, we consider the contributions from \emph{all} pairs, with each scaled by the \emph{similarity} in their labels.

\textbf{Generalized InfoNCE Loss.}
For an input sample $\mathbf{x}$, \simper creates $M$ variant views with different speed labels $\{s_i\}_{i=1}^M$. Given the features of two sets of invariant views $\{\mathbf{z}_i\}_{i=1}^M$, $\{\mathbf{z}'_i\}_{i=1}^M$, we have
\begingroup\makeatletter\def\f@size{8}\check@mathfonts\def\maketag@@@#1{\hbox{\m@th\large\normalfont#1}}
\begin{equation}
\mathcal{L}_{\text{SimPer}}
= \sum_i \ell_{\text{SimPer}}^i
= \sum_i - \sum_{j=1}^M \frac{\exp(w_{i,j})}{\sum_{k=1}^M \exp(w_{i,k})}  \log \frac{\exp (\operatorname{sim}(\mathbf{z}_i, \mathbf{z}'_j) / \nu)}{\sum_{k=1}^M \exp (\operatorname{sim}(\mathbf{z}_i, \mathbf{z}'_k) / \nu)}, \quad w_{i,j} := \operatorname{sim_{\text{label}}}(s_i, s_j),
\label{eqn:simper}
\end{equation}
\endgroup
where $\operatorname{sim}(\cdot)$ denotes the \emph{periodic} feature similarity as described previously, and $\operatorname{sim_{\text{label}}}(\cdot)$ denotes the \emph{continuous} label similarity measure.
In practice, $\operatorname{sim_{\text{label}}}(\cdot)$ can be simply instantiated as inverse of the $L_1$ or $L_2$ label difference (e.g., $1/|s_i - s_j|$).

\textbf{Interpretation.}
$\mathcal{L}_{\text{SimPer}}$ is a simple generalization of the InfoNCE loss from discrete instance discrimination (single target classification) to a weighted loss over all augmented pairs (soft regression variant), where the soft target $\exp(w_{i,j})/ {\sum_k \exp(w_{i,k})}$ is driven by the \emph{label} (speed) similarity $w_{i,j}$ of each pair.
Note that when the label becomes discrete (i.e., $w_{i,j}\in \{0, 1\}$), $\mathcal{L}_{\text{SimPer}}$ degenerates to the original InfoNCE loss.
We demonstrate in Appendix \ref{appendix-subsubsec:gen-con-loss} that such continuity modeling via a generalized loss helps achieve better downstream performance than simply applying InfoNCE.

\vspace{-0.3cm}
\section{Experiments}
\vspace{-0.2cm}
\label{sec:experiment}
\textbf{Datasets.}
We perform extensive experiments on six datasets that span different domains and tasks. Complete descriptions of each dataset are in Appendix \ref{appendix-sec:dataset-details}, Fig.~\ref{appendix-fig:dataset-examples}, and Table \ref{appendix:table:datasets}.

\vspace{-0.2cm}
\begin{Itemize}
    \item \emph{\textbf{RotatingDigits} (Synthetic Dataset)} is a toy periodic learning dataset consists of rotating MNIST digits \citep{deng2012mnist}. The task is to predict the underlying digit rotation frequency.
    \vspace{0.1cm}

    \item \emph{\textbf{SCAMPS} (Human Physiology)} \citep{mcduff2022scamps} consists of 2,800 synthetic videos of avatars with realistic peripheral blood flow. The task is to predict averaged heart rate from input videos.
    \vspace{0.1cm}

    \item \emph{\textbf{UBFC} (Human Physiology)} \citep{bobbia2019unsupervised} contains 42 videos with synchronized gold-standard contact PPG recordings. The task is to predict averaged heart rate from input video clips.


    \item \emph{\textbf{PURE} (Human Physiology)} \citep{stricker2014non} contains 60 videos with synchronized gold-standard contact PPG recordings. The task is to predict averaged heart rate from input video clips.

    \vspace{0.1cm}

    \item \emph{\textbf{Countix} (Action Counting)}. The Countix dataset \citep{dwibedi2020counting} is a subset of the Kinetics \citep{kay2017kinetics} dataset annotated with segments of repeated actions and corresponding counts. The task is to predict the count number given an input video.
    \vspace{0.1cm}

    \item \emph{\textbf{Land Surface Temperature (LST)} (Satellite Sensing)}. LST contains hourly land surface temperature maps over the continental United States for 100 days (April 7th to July 16th, 2022). The task is to predict future temperatures based on past satellite measurements.
\end{Itemize}
\vspace{-0.1cm}

\textbf{Network Architectures.}
We choose a set of logical architectures from prior work for our experiments.
On RotatingDigits and SCAMPS, we employ a simple 3D variant of the CNN architecture as in \citep{yang2022multi}. Following \citep{liu2020multi}, we adopt a variant of TS-CAN model for experiments on UBFC and PURE. Finally, on Countix and LST, we employ ResNet-3D-18 \citep{he2016deep,tran2018resnet3d} as our backbone network. Implementations details are in Appendix \ref{appendix-sec:exp-settings}.

\textbf{Baselines.}
We compare \simper to SOTA SSL methods, including SimCLR \citep{chen2020simple}, MoCo v2 \citep{he2020momentum}, BYOL \citep{grill2020bootstrap}, and CVRL \citep{qian2021spatiotemporal}, as well as a supervised learning counterpart. We provide detailed descriptions in Appendix \ref{appendix-subsec:baselines}.

\textbf{Evaluation Metrics.}
To assess the prediction of continuous targets (e.g., frequency, counts), we use common metrics for regression, such as the mean-average-error (MAE), mean-average-percentage-error (MAPE), Pearson correlation ($\rho$), and error Geometric Mean (GM) \citep{yang2021delving}.

\begin{table}[t]
\setlength{\tabcolsep}{2.5pt}
\parbox{.49\linewidth}{
\caption{\small Feature evaluation results on RotatingDigits.}
\vspace{-1.5pt}
\label{exp:table:rotatingdigits}
\small
\begin{center}
\adjustbox{max width=0.485\textwidth}{
\begin{tabular}{lcccc}
\toprule[1.5pt]
 & \multicolumn{2}{c}{FFT} & \multicolumn{2}{c}{1-NN} \\
\cmidrule(lr){2-3} \cmidrule(lr){4-5}
Metrics & MAE$^\downarrow$ & MAPE$^\downarrow$ & MAE$^\downarrow$ & MAPE$^\downarrow$ \\ \midrule\midrule
\textsc{SimCLR}~\citep{chen2020simple} & 2.96 & 109.27 & 0.98 & 48.30 \\[1.2pt]
\textsc{MoCo v2}~\citep{he2020momentum} & 2.83 & 90.78 & 0.62 & 32.74 \\[1.2pt]
\textsc{BYOL}~\citep{grill2020bootstrap} & 2.20 & 78.43 & 0.46 & 22.08 \\[1.2pt]
\textsc{CVRL}~\citep{qian2021spatiotemporal} & 1.69 & 49.09 & 0.38 & 14.41 \\[1.2pt]
\grayrow
\textbf{\textsc{SimPer}} & \textbf{0.22} & \textbf{16.49} & \textbf{0.09} & \textbf{4.51} \\
\midrule
\textsc{Gains} & \textcolor{darkgreen}{\texttt{+}\textbf{1.47}} & \textcolor{darkgreen}{\texttt{+}\textbf{32.60}} & \textcolor{darkgreen}{\texttt{+}\textbf{0.29}} & \textcolor{darkgreen}{\texttt{+}\textbf{9.90}} \\
\bottomrule[1.5pt]
\end{tabular}}
\end{center}
}
\parbox{.5\linewidth}{
\caption{\small Feature evaluation results on SCAMPS.}
\label{exp:table:scamps}
\small
\begin{center}
\adjustbox{max width=0.491\textwidth}{
\begin{tabular}{lcccc}
\toprule[1.5pt]
 & \multicolumn{2}{c}{FFT} & \multicolumn{2}{c}{1-NN} \\
\cmidrule(lr){2-3} \cmidrule(lr){4-5}
Metrics & MAE$^\downarrow$ & MAPE$^\downarrow$ & MAE$^\downarrow$ & MAPE$^\downarrow$ \\ \midrule\midrule
\textsc{SimCLR}~\citep{chen2020simple} & 27.48 & 38.39 & 34.09 & 40.79 \\[1.2pt]
\textsc{MoCo v2}~\citep{he2020momentum} & 28.16 & 40.23 & 35.61 & 42.47 \\[1.2pt]
\textsc{BYOL}~\citep{grill2020bootstrap} & 26.15  & 37.34 & 32.77  & 38.26 \\[1.2pt]
\textsc{CVRL}~\citep{qian2021spatiotemporal} & 27.67 & 38.80 & 33.32 & 39.54 \\[1.2pt]
\grayrow
\textbf{\textsc{SimPer}} & \textbf{14.45} & \textbf{22.09} & \textbf{13.75} & \textbf{18.64} \\
\midrule
\textsc{Gains} & \textcolor{darkgreen}{\texttt{+}\textbf{11.70}} & \textcolor{darkgreen}{\texttt{+}\textbf{15.25}} & \textcolor{darkgreen}{\texttt{+}\textbf{19.02}} & \textcolor{darkgreen}{\texttt{+}\textbf{19.62}} \\
\bottomrule[1.5pt]
\end{tabular}}
\end{center}
}
\vspace{-0.6cm}
\end{table}

\begin{table}[t]
\setlength{\tabcolsep}{2.5pt}
\parbox{.49\linewidth}{
\caption{\small Feature evaluation results on UBFC.}
\label{exp:table:ubfc}
\small
\begin{center}
\adjustbox{max width=0.485\textwidth}{
\begin{tabular}{lcccc}
\toprule[1.5pt]
 & \multicolumn{2}{c}{FFT} & \multicolumn{2}{c}{1-NN} \\
\cmidrule(lr){2-3} \cmidrule(lr){4-5}
Metrics & MAE$^\downarrow$ & MAPE$^\downarrow$ & MAE$^\downarrow$ & MAPE$^\downarrow$ \\ \midrule\midrule
\textsc{SimCLR}~\citep{chen2020simple} & 16.92 & 14.73 & 16.23 & 18.62 \\[1.2pt]
\textsc{MoCo v2}~\citep{he2020momentum} & 14.64 & 13.17 & 15.12 & 16.56 \\[1.2pt]
\textsc{BYOL}~\citep{grill2020bootstrap} & 17.86 & 16.90 & 18.13 & 19.34 \\[1.2pt]
\textsc{CVRL}~\citep{qian2021spatiotemporal} & 11.75 & 10.67 & 12.36 & 13.38 \\[1.2pt]
\grayrow
\textbf{\textsc{SimPer}} & \textbf{8.78} & \textbf{7.46} & \textbf{8.92} & \textbf{10.21} \\
\midrule
\textsc{Gains} & \textcolor{darkgreen}{\texttt{+}\textbf{2.97}} & \textcolor{darkgreen}{\texttt{+}\textbf{3.21}} & \textcolor{darkgreen}{\texttt{+}\textbf{3.44}} & \textcolor{darkgreen}{\texttt{+}\textbf{3.17}} \\
\bottomrule[1.5pt]
\end{tabular}}
\end{center}
}
\parbox{.49\linewidth}{
\caption{\small Feature evaluation results on PURE.}
\label{exp:table:pure}
\small
\begin{center}
\adjustbox{max width=0.485\textwidth}{
\begin{tabular}{lcccc}
\toprule[1.5pt]
 & \multicolumn{2}{c}{FFT} & \multicolumn{2}{c}{1-NN} \\
\cmidrule(lr){2-3} \cmidrule(lr){4-5}
Metrics & MAE$^\downarrow$ & MAPE$^\downarrow$ & MAE$^\downarrow$ & MAPE$^\downarrow$ \\ \midrule\midrule
\textsc{SimCLR}~\citep{chen2020simple} & 23.70 & 22.07 & 29.48 & 31.44 \\[1.2pt]
\textsc{MoCo v2}~\citep{he2020momentum} & 24.23 & 24.08 & 30.82 & 33.95 \\[1.2pt]
\textsc{BYOL}~\citep{grill2020bootstrap} & 23.24  & 21.78 & 29.27 & 31.03 \\[1.2pt]
\textsc{CVRL}~\citep{qian2021spatiotemporal} & 19.27 & 18.94 & 22.08 & 23.75 \\[1.2pt]
\grayrow
\textbf{\textsc{SimPer}} & \textbf{13.97} & \textbf{12.88} & \textbf{14.03} & \textbf{15.35} \\
\midrule
\textsc{Gains} & \textcolor{darkgreen}{\texttt{+}\textbf{5.30}} & \textcolor{darkgreen}{\texttt{+}\textbf{6.06}} & \textcolor{darkgreen}{\texttt{+}\textbf{8.05}} & \textcolor{darkgreen}{\texttt{+}\textbf{8.40}} \\
\bottomrule[1.5pt]
\end{tabular}}
\end{center}
}
\vspace{-0.6cm}
\end{table}

\begin{table}[!t]
\setlength{\tabcolsep}{4pt}
\parbox{.52\linewidth}{
\caption{\small Feature evaluation results on Countix.}
\label{exp:table:countix}
\small
\begin{center}
\adjustbox{max width=0.505\textwidth}{
\begin{tabular}{lcccc}
\toprule[1.5pt]
 & \multicolumn{2}{c}{FFT} & \multicolumn{2}{c}{1-NN} \\
\cmidrule(lr){2-3} \cmidrule(lr){4-5}
Metrics & MAE$^\downarrow$ & GM$^\downarrow$ & MAE$^\downarrow$ & GM$^\downarrow$ \\ \midrule\midrule
\textsc{SimCLR}~\citep{chen2020simple} & 3.90 & 2.26 & 4.43 & 3.19 \\[1.2pt]
\textsc{MoCo v2}~\citep{he2020momentum} & 3.75 & 2.18 & 3.96 & 3.04 \\[1.2pt]
\textsc{BYOL}~\citep{grill2020bootstrap} & 3.26 & 1.87 & 3.72 & 2.66 \\[1.2pt]
\textsc{CVRL}~\citep{qian2021spatiotemporal} & 2.81 & 1.38 & 3.15 & 2.12 \\[1.2pt]
\grayrow
\textbf{\textsc{SimPer}} & \textbf{2.06} & \textbf{0.98} & \textbf{2.76} & \textbf{1.84} \\
\midrule
\textsc{Gains} & \textcolor{darkgreen}{\texttt{+}\textbf{0.75}} & \textcolor{darkgreen}{\texttt{+}\textbf{0.40}} & \textcolor{darkgreen}{\texttt{+}\textbf{0.99}} & \textcolor{darkgreen}{\texttt{+}\textbf{0.28}} \\
\bottomrule[1.5pt]
\end{tabular}}
\end{center}
}
\parbox{.46\linewidth}{
\caption{\small Feature evaluation results on LST.}
\label{exp:table:lst}
\small
\begin{center}
\adjustbox{max width=0.455\textwidth}{
\begin{tabular}{lccc}
\toprule[1.5pt]
 & \multicolumn{3}{c}{Linear Probing} \\
\cmidrule(lr){2-4}
Metrics & MAE$^\downarrow$ & MAPE$^\downarrow$ & $\rho^\uparrow$ \\ \midrule\midrule
\textsc{SimCLR}~\citep{chen2020simple} & 5.12 & 0.20 & 0.89 \\[1.2pt]
\textsc{MoCo v2}~\citep{he2020momentum} & 5.16 & 0.20 & 0.89 \\[1.2pt]
\textsc{BYOL}~\citep{grill2020bootstrap} & 5.71 & 0.24 & 0.86 \\[1.2pt]
\textsc{CVRL}~\citep{qian2021spatiotemporal} & 4.88 & \textbf{0.18} & \textbf{0.91} \\[1.2pt]
\grayrow
\textbf{\textsc{SimPer}} & \textbf{4.84} & \textbf{0.18} & 0.90 \\
\midrule
\textsc{Gains} & \textcolor{darkgreen}{\texttt{+}\textbf{0.04}} & \textcolor{darkgreen}{\texttt{+}\textbf{0.00}} & \textcolor{lightblue}{\texttt{-}\textbf{0.01}} \\
\bottomrule[1.5pt]
\end{tabular}}
\end{center}
}
\vspace{-0.3cm}
\end{table}

\begin{table}[!t]
\setlength{\tabcolsep}{3pt}
\caption{\small 
Fine-tune evaluation results on all datasets. We first pre-train the feature encoder using different SSL methods, then fine-tune the whole network initialized with the pre-trained weights.
}
\vspace{-8pt}
\label{exp:table:finetune}
\small
\begin{center}
\adjustbox{max width=\textwidth}{
\begin{tabular}{lcccccccccccc}
\toprule[1.5pt]
 & \multicolumn{2}{c}{RotatingDigits} & \multicolumn{2}{c}{SCAMPS} & \multicolumn{2}{c}{UBFC} & \multicolumn{2}{c}{PURE} & \multicolumn{2}{c}{Countix} & \multicolumn{2}{c}{LST} \\
\cmidrule(lr){2-3} \cmidrule(lr){4-5} \cmidrule(lr){6-7} \cmidrule(lr){8-9} \cmidrule(lr){10-11} \cmidrule(lr){12-13}
Metrics & MAE$^\downarrow$ & MAPE$^\downarrow$ & MAE$^\downarrow$ & MAPE$^\downarrow$ & MAE$^\downarrow$ & MAPE$^\downarrow$ & MAE$^\downarrow$ & MAPE$^\downarrow$ & MAE$^\downarrow$ & GM$^\downarrow$ & MAE$^\downarrow$ & $\rho^\uparrow$ \\ \midrule\midrule
\textsc{Supervised} & 0.72 & 28.96 & 3.61 & 5.33 & 5.13 & 4.72 & 4.25 & 4.93 & 1.50 & 0.73 & 1.54 & \textbf{0.96} \\ \midrule
\textsc{SimCLR}~\citep{chen2020simple} & 0.69 & 26.54 & 4.96 & 6.92 & 5.32 & 4.96 & 4.86 & 5.32 & 1.58 & 0.80 & 1.54 & 0.95 \\[1.2pt]
\textsc{MoCo v2}~\citep{he2020momentum} & 0.64 & 24.73  & 5.33 & 7.24 & 5.05 & 4.64 & 4.97 & 5.60 & 1.54 & 0.79 & 1.53 & 0.95 \\[1.2pt]
\textsc{BYOL}~\citep{grill2020bootstrap} & 0.39 & 20.91 & 3.49 & 5.27 & 5.51 & 5.07 & 4.28 & 4.97 & 1.47 & 0.71 & 1.62 & 0.92 \\[1.2pt]
\textsc{CVRL}~\citep{qian2021spatiotemporal} & 0.34 & 18.82 & 5.52 & 7.34 & 5.07 & 4.70 & 4.19 & 4.71 & 1.48 & 0.71 & 1.49 & \textbf{0.96} \\[1.2pt]
\grayrow
\textbf{\textsc{SimPer}} & \textbf{0.20} & \textbf{14.33} & \textbf{3.27} & \textbf{4.89} & \textbf{4.24} & \textbf{3.97} & \textbf{3.89} & \textbf{4.01} & \textbf{1.33} & \textbf{0.59} & \textbf{1.47} & \textbf{0.96} \\
\midrule
\textsc{Gains vs. Supervised} & \textcolor{darkgreen}{\texttt{+}\textbf{0.52}} & \textcolor{darkgreen}{\texttt{+}\textbf{14.63}} & \textcolor{darkgreen}{\texttt{+}\textbf{0.34}} & \textcolor{darkgreen}{\texttt{+}\textbf{0.44}} & \textcolor{darkgreen}{\texttt{+}\textbf{0.89}} & \textcolor{darkgreen}{\texttt{+}\textbf{0.75}} & \textcolor{darkgreen}{\texttt{+}\textbf{0.36}} & \textcolor{darkgreen}{\texttt{+}\textbf{0.92}} & \textcolor{darkgreen}{\texttt{+}\textbf{0.17}} & \textcolor{darkgreen}{\texttt{+}\textbf{0.14}} & \textcolor{darkgreen}{\texttt{+}\textbf{0.07}} & \textcolor{darkgreen}{\texttt{+}\textbf{0.00}} \\
\bottomrule[1.5pt]
\end{tabular}}
\end{center}
\vspace{-0.2cm}
\end{table}

\vspace{-0.2cm}
\subsection{Main Results}
\vspace{-0.15cm}
\label{subsec:exp-main}

We report the main results in this section for all datasets. Complete training details, hyper-parameter settings, and additional results are provided in Appendix \ref{appendix-sec:exp-settings} and \ref{appendix-sec:additional-results}.

\textbf{Feature Evaluation.}
Following the literature \citep{he2020momentum, chen2020simple}, we first evaluate the representations learned by different methods.
For dense prediction task (e.g., LST), we use the \emph{linear probing} protocol by training a linear regressor on top of the fixed features.
For tasks whose targets are frequency information, we directly evaluate the learned features using a Fourier transform (\textbf{FFT}) and a nearest neighbor classifier (\textbf{1-NN}).
Table~\ref{exp:table:rotatingdigits}, \ref{exp:table:scamps}, \ref{exp:table:ubfc}, \ref{exp:table:pure}, \ref{exp:table:countix}, \ref{exp:table:lst} show the feature evaluation results of \simper compared to SOTA SSL methods.
As the tables confirm, across different datasets with various common tasks, \simper is able to learn better representations that achieve the best performance. Furthermore, in certain datasets, the relative improvements are even larger than $50\%$.

\textbf{Fine-tuning.}
Practically, to harness the power of pre-trained representations, fine-tuning the whole network with the encoder initialized using pre-trained weights is a widely adopted approach \citep{he2020momentum}.
To evaluate whether \simper pre-training is helpful for each downstream task, we fine-tune the whole network and compare the final performance. The details of the setup for each dataset and algorithm can be found in Appendix~\ref{appendix-sec:exp-settings}.
As Table~\ref{exp:table:finetune} confirms, across different datasets, \simper consistently outperforms all other SOTA SSL methods, and obtains better results compared to the supervised baseline.
This demonstrates that \simper is able to capture meaningful periodic information that is beneficial to the downstream tasks.

\begin{figure*}[t]
\centering
\subfigure[Learned representations under different training dataset sizes]{
    \label{fig:exp-dataset-size:qualitative}
    \includegraphics[height=0.41\textwidth]{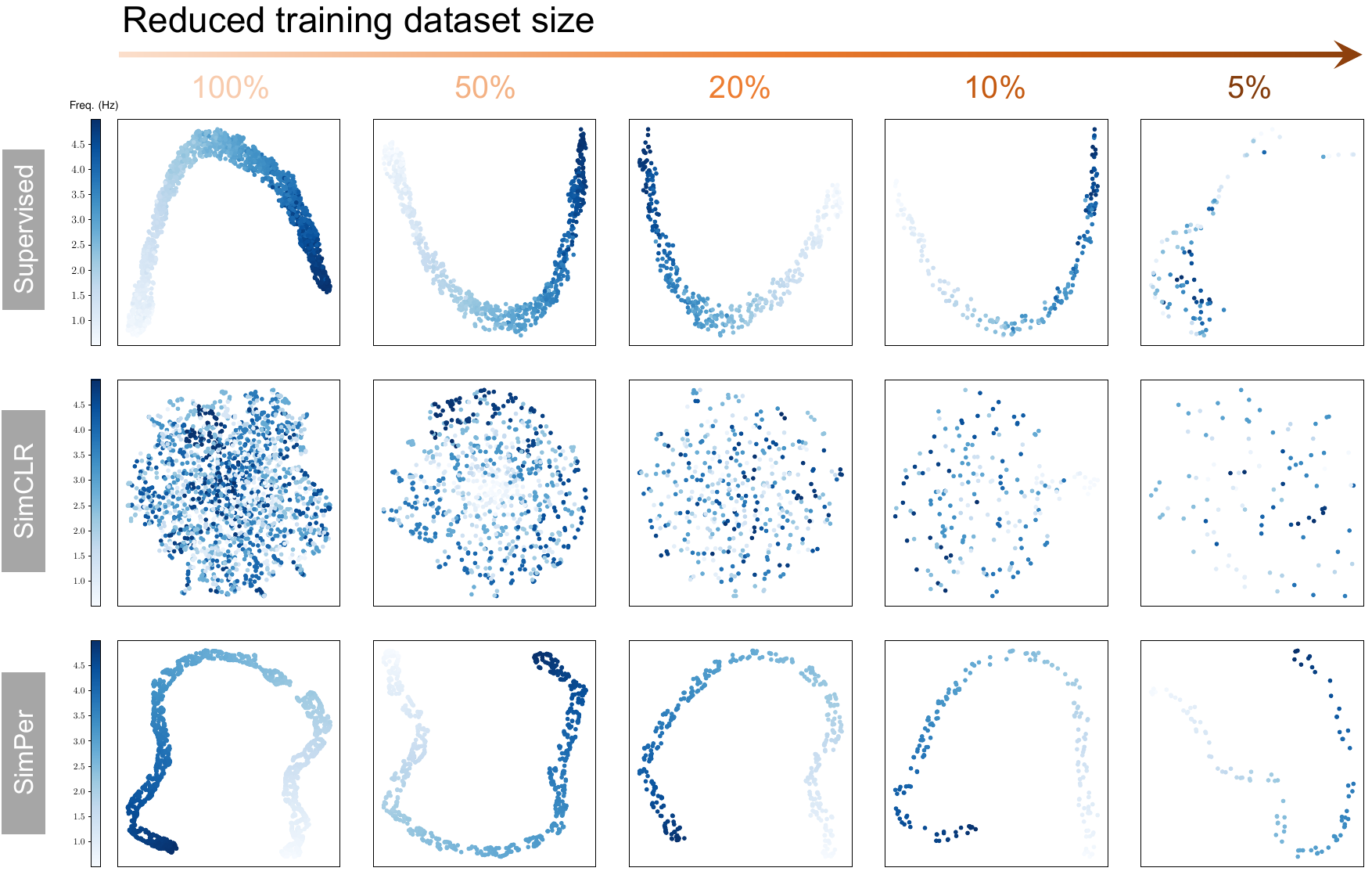}
}
\hspace{1ex}
\subfigure[Quantitative error changes]{
    \label{fig:exp-dataset-size:quantitative}
    \includegraphics[height=0.41\textwidth]{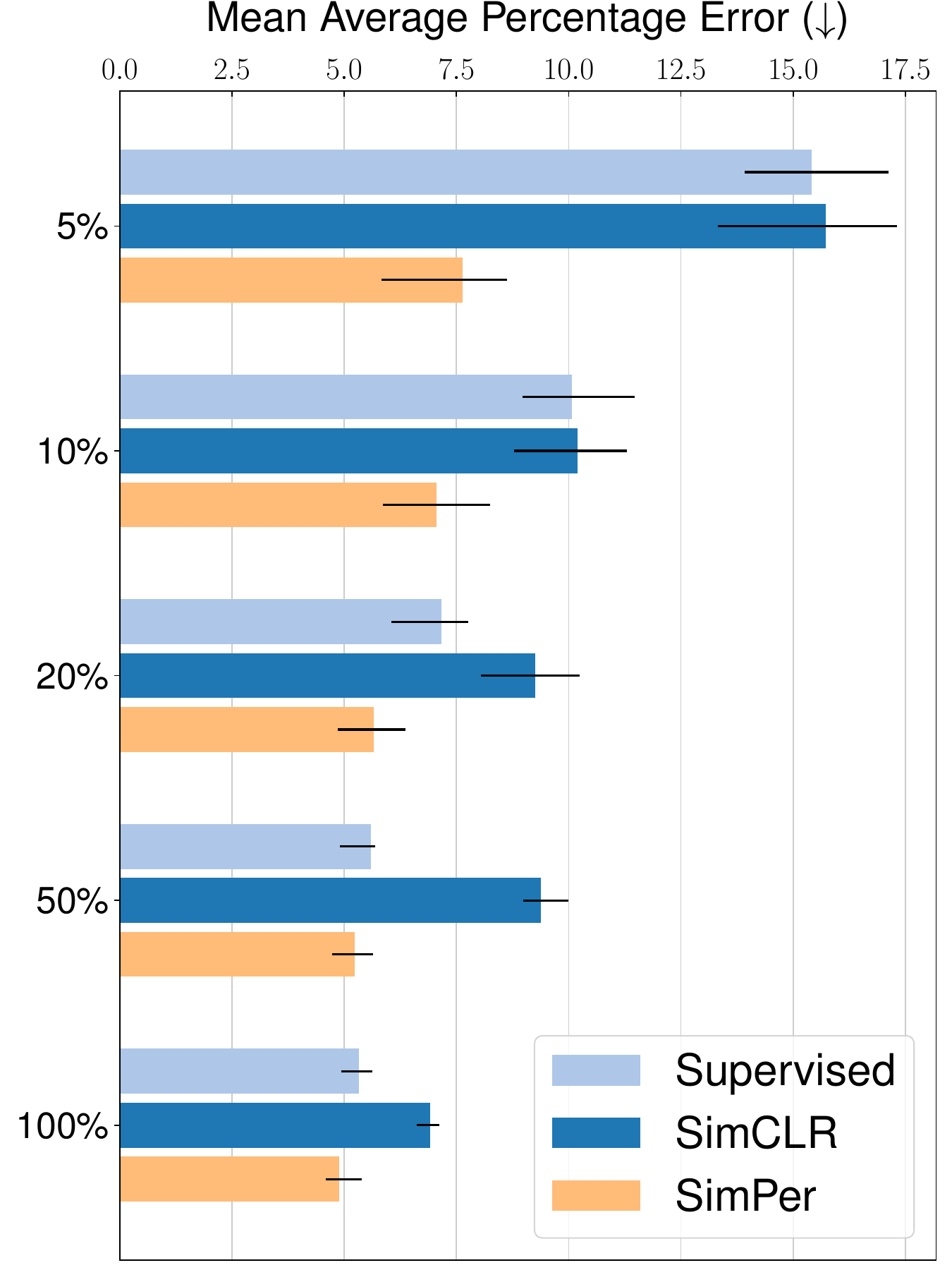}
}
\vspace{-0.3cm}
\caption{\small 
\textbf{Data efficiency analysis of \simper}. \textbf{(a)} Learned representations of different algorithms on RotatingDigits when training dataset size reduces from $100\%$ to $5\%$. \textbf{(b)} The quantitative MAPE errors on SCAMPS with varying training dataset sizes.
Complete quantitative results are provided in Appendix~\ref{appendix-subsec:data-efficiency-reduced-training-data}.
}
\label{fig:exp-dataset-size}
\vspace{-0.2cm}
\end{figure*}


\vspace{-0.2cm}
\subsection{Data Efficiency}
\vspace{-0.15cm}
\label{subsec:exp-data-efficiency}

In real-world periodic learning applications, data is often prohibitively expensive to obtain.
To study the data efficiency of \simper, we manually reduce the overall size of RotatingDigits, and plot the representations learned as well as the final fine-tuning accuracy of different methods in Fig.~\ref{fig:exp-dataset-size}.

As the figure confirms, when the dataset size is large (e.g., using $100\%$ of the data), both supervised learning baseline and \simper can learn good representations (Fig.~\ref{fig:exp-dataset-size:qualitative}) and achieve low test errors (Fig.~\ref{fig:exp-dataset-size:quantitative}).
However, when the training dataset size becomes smaller, the learned representations using supervised learning get worse, and eventually lose the frequency information and resolution when only $5\%$ of the data is available. Correspondingly, the final error in this extreme case also becomes much higher.
In contrast, even with small number of training data, \simper can consistently learn the periodic information and maintain high frequency resolution, with significant performance gains especially when the available data amount is small.

\setlength\intextsep{-5pt}
\begin{wraptable}[9]{r}{0.45\textwidth}
\setlength{\tabcolsep}{6pt}
\caption{\small Transfer learning results.}
\vspace{-10pt}
\label{table:transfer-results}
\small
\begin{center}
\adjustbox{max width=0.45\textwidth}{
\begin{tabular}{lcccc}
\toprule[1.5pt]
 & \multicolumn{2}{c}{UBFC $\rightarrow$ PURE}       & \multicolumn{2}{c}{PURE $\rightarrow$ UBFC} \\
\cmidrule(lr){2-3} \cmidrule(lr){4-5}
Metrics & MAE$^\downarrow$ & MAPE$^\downarrow$ & MAE$^\downarrow$ & MAPE$^\downarrow$ \\ \midrule\midrule
\textsc{Supervised} & 7.83 & 8.85 & 3.15 & 3.11 \\[1.2pt]
\textsc{SimCLR} & 7.86 & 8.79 & 3.46 & 3.80 \\[1.2pt]
\grayrow
\textbf{\textsc{SimPer}} & \textbf{6.46} & \textbf{6.98} & \textbf{2.76} & \textbf{2.38} \\ \midrule
\textsc{Gains} & \textcolor{darkgreen}{\texttt{+}\textbf{1.37}} & \textcolor{darkgreen}{\texttt{+}\textbf{1.87}} & \textcolor{darkgreen}{\texttt{+}\textbf{0.39}} & \textcolor{darkgreen}{\texttt{+}\textbf{0.73}} \\
\bottomrule[1.5pt]
\end{tabular}}
\end{center}
\end{wraptable}

\vspace{-0.2cm}
\subsection{Transfer Learning}
\vspace{-0.15cm}
\label{subsec:exp-transfer-learning}

We evaluate whether the self-supervised representations are transferable across datasets. We use UBFC and PURE, which share the same prediction task.
Following \citep{chen2020simple}, we fine-tune the pre-trained model on the new dataset, and compare the performance across both SSL and supervised methods.
Table \ref{table:transfer-results} reports the results, where in both cases, \simper is able to achieve better final performance compared to supervised and SSL baselines, showing its ability to learn transferable periodic representations across different datasets.

\begin{figure}[!t]
\begin{center}
\includegraphics[width=\linewidth]{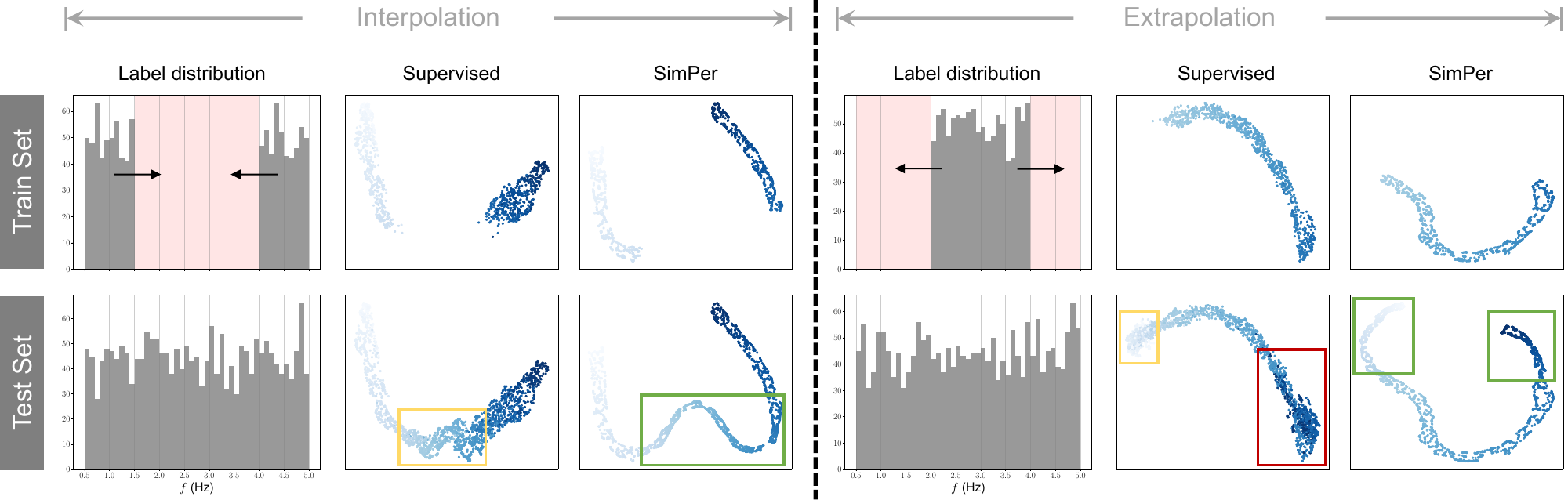}
\end{center}
\vspace{-0.4cm}
\caption{\small \textbf{Zero-shot generalization analysis.} We create training sets with missing target frequencies and keep test sets evenly distributed across the target range. \textcolor{darkgreen}{Green} regions indicate successful generalization with high frequency resolution. \textcolor{medshot}{Yellow} regions indicate successful generalization but with low frequency resolution. \textcolor{darkred}{Red} regions represent failed generalization. \simper learns robust representations that generalize to unseen targets.}
\label{fig:exp-zero-shot}
\vspace{-0.3cm}
\end{figure}

\vspace{-0.2cm}
\subsection{Zero-shot Generalization to Unseen Targets}
\vspace{-0.15cm}
\label{subsec:exp-zero-shot}

Given the continuous nature of the frequency domain, periodic learning tasks can (and almost certainly will) have unseen frequency targets during training, which motivates the need for target (frequency) extrapolation and interpolation.
To investigate \textbf{zero-shot} generalization to unseen targets, we manually create training sets that have certain missing targets (Fig.~\ref{fig:exp-zero-shot}), while making the test sets evenly distributed across the target range.
As Fig.~\ref{fig:exp-zero-shot} confirms, in the interpolation case, both supervised learning and \simper can successfully interpolate the missing targets. However, the quality of interpolation varies: For supervised learning, the frequency resolution is low within the interpolation range, resulting in mixed representations for a wide missing range. In contrast, \simper learns better representations with higher frequency resolution, which has desirable discriminative properties.

\setlength\intextsep{-4pt}
\begin{wraptable}[10]{r}{0.45\textwidth}
\setlength{\tabcolsep}{6pt}
\caption{\small Mean absolute error (MAE, $\downarrow$) results for zero-shot generalization analysis.}
\vspace{-10pt}
\label{table:zero-shot}
\small
\begin{center}
\adjustbox{max width=0.45\textwidth}{
\begin{tabular}{lcccc}
\toprule[1.5pt]
 & \multicolumn{2}{c}{\textbf{Interpolation}} & \multicolumn{2}{c}{\textbf{Extrapolation}} \\
\cmidrule(lr){2-3} \cmidrule(lr){4-5}
 & Seen & \graycell{Unseen} & Seen & \graycell{Unseen} \\ \midrule\midrule
\textsc{Supervised} & {0.09} & \graycell{0.85} & {0.03} & \graycell{1.74} \\[1.2pt]
\textbf{\textsc{SimPer}} & \textbf{0.05} & \graycell{\textbf{0.07}} & \textbf{0.02} & \graycell{\textbf{0.02}} \\ \midrule
\textsc{Gains} & \textcolor{darkgreen}{\texttt{+}\textbf{0.04}} & \graycell{\textcolor{darkgreen}{\texttt{+}\textbf{0.78}}} & \textcolor{darkgreen}{\texttt{+}\textbf{0.01}} & \graycell{\textcolor{darkgreen}{\texttt{+}\textbf{1.72}}} \\
\bottomrule[1.5pt]
\end{tabular}}
\end{center}
\end{wraptable}

Furthermore, in the extrapolation case, in the lower frequency range, both methods extrapolate reasonably well, with \simper capturing a higher frequency resolution. However, when extrapolating to a higher frequency range, the supervised baseline completely fails to generalize, with learned features largely overlapping with the existing frequency targets in the training set. In contrast, \simper is able to generalize robustly even for the higher unseen frequency range, demonstrating its effectiveness of generalization to distribution shifts and unseen targets.
Quantitative results in Table \ref{table:zero-shot} confirm the observations.

\vspace{-0.2cm}
\subsection{Robustness to Spurious Correlations}
\vspace{-0.15cm}
\label{subsec:exp-spurious-corr}

We show that \simper is able to deal with spurious correlations that arise in data, while existing SSL methods often fail to learn generalizable features.
Specifically, RotatingDigits dataset naturally has a spurious target: the digit appearance (number). We further enforce this information by coloring different digits with different colors as in \citep{arjovsky2019irm}. We then construct a spuriously correlated training set by assigning a unique rotating frequency range to a specific digit, i.e., $[0.5\text{Hz},$ $1\text{Hz}]$ for digit $0$, $[1\text{Hz}, 1.5\text{Hz}]$ for digit $1$, etc, while removing the spurious correlations in test set.

\begin{figure}[!t]
\begin{center}
\includegraphics[width=0.85\linewidth]{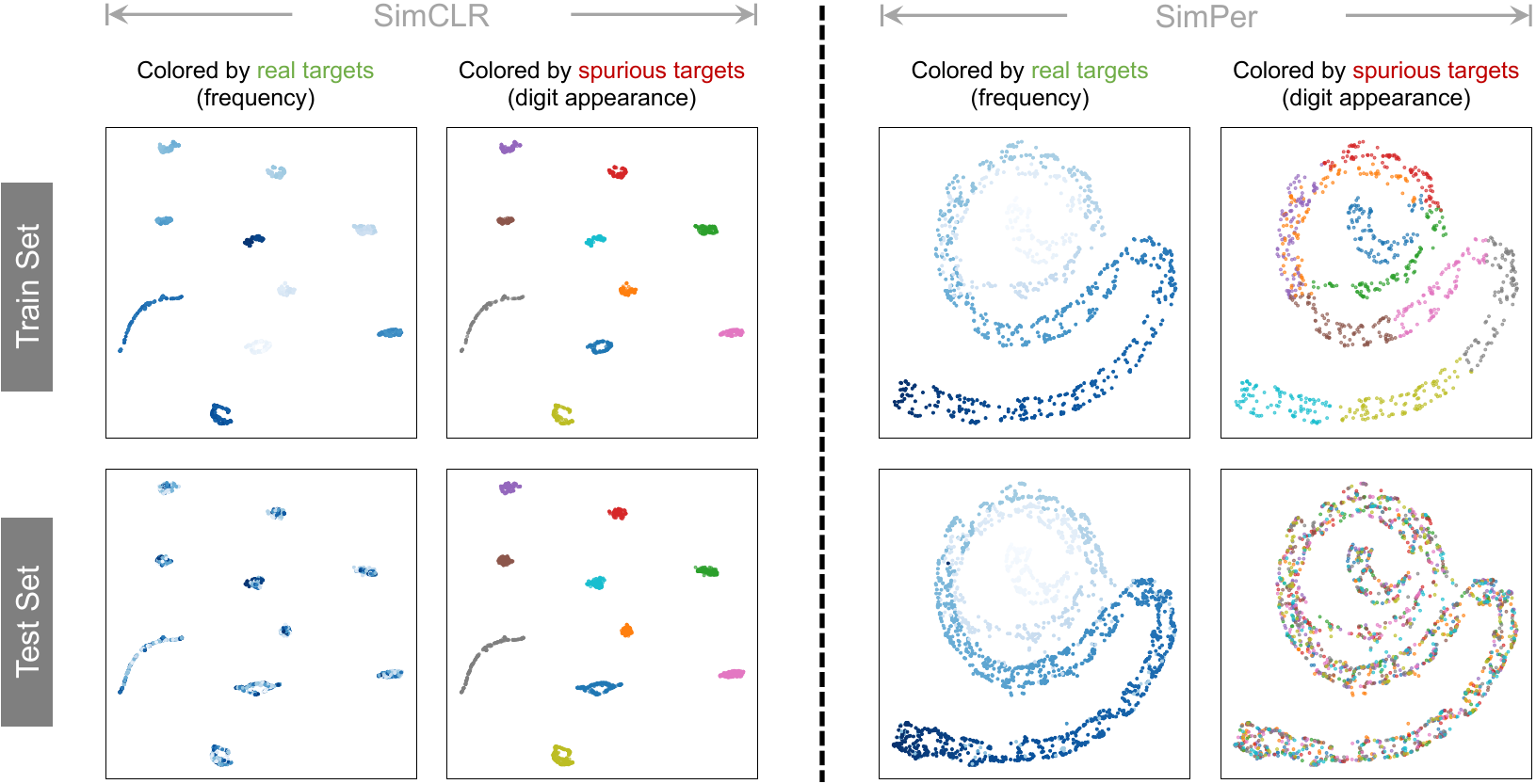}
\end{center}
\vspace{-0.4cm}
\caption{\small 
\textbf{Robustness to spurious correlations.} We make the target frequency spuriously correlated with digit appearances in training set, while removing this correspondence in test set.
\simper is able to capture underlying periodic information \& learn robust representations that generalize.
Quantitative results are in Appendix~\ref{appendix-subsec:spurious}.}
\label{fig:spurious}
\vspace{-0.4cm}
\end{figure}

As Fig.~\ref{fig:spurious} verifies, SimCLR is easy to learn information that is spuriously correlated in the training data, but not the actual target of interest (frequency). As a result, the learned representations do not generalize.
In contrast, \simper learns the underlying frequency information even in the presence of strong spurious correlations, demonstrating its ability to learn robust representations that generalize.

\vspace{-0.2cm}
\subsection{Further Analysis and Ablation Studies}
\vspace{-0.15cm}
\label{subsec:exp-ablation}

\textbf{Amount of labeled data for fine-tuning (Appendix~\ref{appendix-subsec:data-efficiency-amount-labeled-data}).}
We show that when the amount of labeled data is limited for fine-tuning, \simper still substantially outperforms baselines by a large margin, achieving a $67\%$ relative improvement in MAE even when the labeled data fraction is only $5\%$.

\textbf{Ablation: Frequency augmentation range (Appendix~\ref{appendix-subsubsec:ablation-freq-range}).}
We study the effects of different speed (frequency) augmentation ranges when creating periodicity-variant views (Table \ref{appendix:table:ablation-freq-range}). While a proper range can lead to certain gains, \simper is reasonably robust to different choices.

\textbf{Ablation: Number of augmented views (Appendix~\ref{appendix-subsubsec:ablation-num-augs}).}
We investigate the influence of different number of augmented views (i.e., $M$) in \simper. Interestingly, we find \simper is surprisingly robust to different $M$ in a given range (Table \ref{appendix:table:ablation-num-views}), where larger $M$ often delivers better results.

\textbf{Ablation: Choices of different similarity metrics (Appendix~\ref{appendix-subsubsec:similarity-metrics}).}
We explore the effects of different periodic similarity measures in \simper, where we show that \simper is robust to all aforementioned periodic similarity measures, achieving similar performances (Table \ref{appendix:table:ablation-similarity-metrics}).

\textbf{Ablation: Effectiveness of generalized contrastive loss (Appendix~\ref{appendix-subsubsec:gen-con-loss}).}
We confirm the effectiveness of the generalized contrastive loss by showing its consistent performance gains across all six datasets, as compared to the vanilla InfoNCE loss (Table \ref{appendix:table:ablation-gen-con-loss}).



\vspace{-0.25cm}
\section{Conclusion}
\vspace{-0.15cm}
\label{sec:conclusion}
We present \simper, a simple and effective SSL framework for learning periodic information from data. \simper develops customized periodicity-variant and invariant augmentations, periodic feature similarity, and a generalized contrastive loss to exploit periodic inductive biases.
Extensive experiments on different datasets over various real-world applications verify the superior performance of \simper, highlighting its intriguing properties such as better efficiency, robustness \& generalization.

\bibliography{iclr2023}
\bibliographystyle{iclr2023}

\newpage
\appendix

\section{Pseudo Code for SimPer}
\label{appendix-sec:pseudo-code}

We provide the pseudo code of \simper in Algorithm~\ref{alg:simper}.
\vspace{0.5cm}
\begin{algorithm}[h]
   \caption{Simple Self-Supervised Learning of Periodic Targets (\simper)}
   \label{alg:simper}
\begin{algorithmic}
   \STATE {\bfseries Input:} Unlabeled training set $\mathcal{D}=\{ (\mathbf{x}_i) \}_{i=1}^{N}$, total training epochs $E$, periodicity-variant augme-ntations $\tau(\cdot)$, periodicity-invariant augmentations $\sigma(\cdot)$, number of variant views $M$, encoder $f$
   \FOR{$e = 0$ \textbf{to} $E$}
   \REPEAT
   \STATE Sample a mini-batch $\{ (\mathbf{x}^{(l)}) \}_{l=1}^n$ from $\mathcal{D}$
   \FOR{$l=1$ \textbf{to} $n$ (in parallel)}
   \STATE $\{\mathbf{x}^{(l)}_i\}_{i=1}^M \leftarrow \tau\left(\mathbf{x}^{(l)}\right)$ \hfill \texttt{//}~~\texttt{M}~~\texttt{variant}~~\texttt{views}~~\texttt{for}~~\texttt{$l$-th}~~\texttt{sample}
   \STATE $\{\mathbf{x}^{(l)}_i\}_{i=1}^M,\  \{{\mathbf{x}'_i}^{(l)}\}_{i=1}^M \leftarrow \sigma \left(\{\mathbf{x}^{(l)}_i\}_{i=1}^M \right)$ \hfill \texttt{//}~~\texttt{two}~~\texttt{sets}~~\texttt{of}~~\texttt{invariant}~~\texttt{views}
   \STATE $\{\mathbf{z}^{(l)}_i\}_{i=1}^M,\  \{{\mathbf{z}'_i}^{(l)}\}_{i=1}^M \leftarrow f \left(\{\mathbf{x}^{(l)}_i\}_{i=1}^M \right),\  f \left(\{{\mathbf{x}'_i}^{(l)}\}_{i=1}^M \right)$
   \STATE Calculate $\ell_\text{SimPer}^{(l)}$ for $l$-th sample using $\{\mathbf{z}^{(l)}_i\}_{i=1}^M,\  \{{\mathbf{z}'_i}^{(l)}\}_{i=1}^M$ based on Eqn.~(\ref{eqn:simper})
   \ENDFOR
   \STATE Calculate $\mathcal{L}_{\text{SimPer}}$ using $\frac{1}{n} \sum_{l=1}^n \ell_\text{SimPer}^{(l)}$ and do one training step
   \UNTIL{iterate over all training samples at current epoch $e$}
   \ENDFOR
\end{algorithmic}
\end{algorithm}
\vspace{0.4cm}

\begin{figure*}[t]
\centering
\subfigure[RotatingDigits]{
    \label{appendix-subfig:rotatingdigits}
    \includegraphics[height=0.41\textwidth]{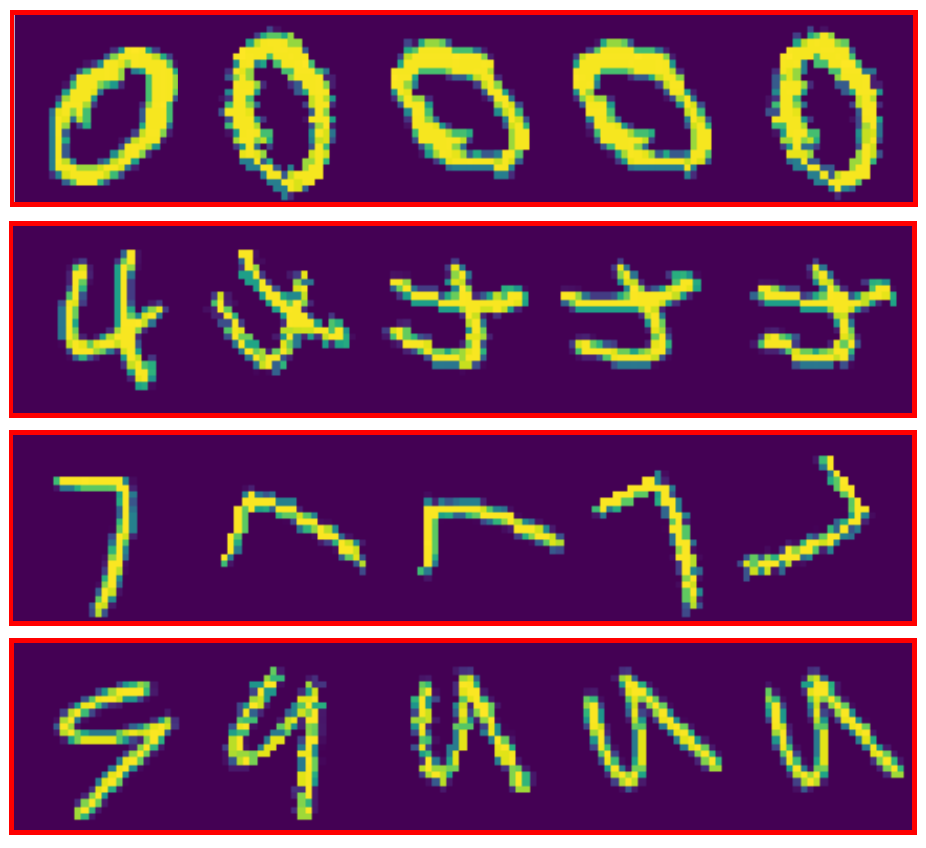}
}
\hfill
\subfigure[SCAMPS~\citep{mcduff2022scamps}]{
    \label{appendix-subfig:scamps}
    \includegraphics[height=0.41\textwidth]{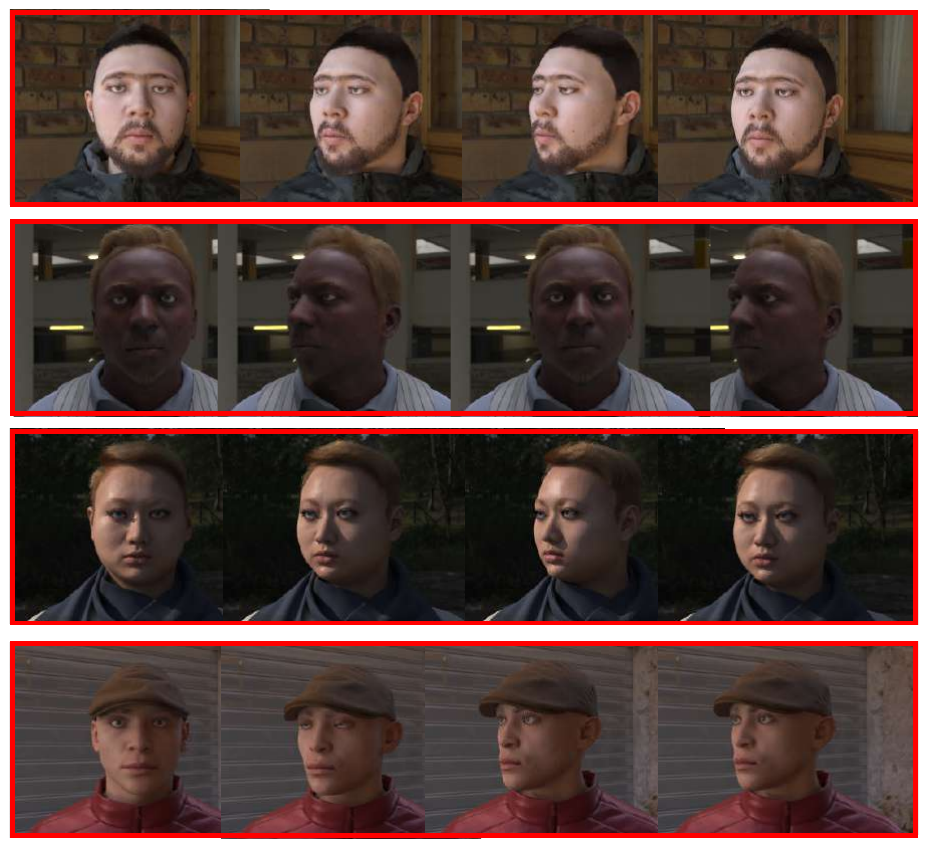}
}
\\
\subfigure[Land Surface Temperature (LST)]{
    \label{appendix-subfig:lst}
    \includegraphics[height=0.41\textwidth]{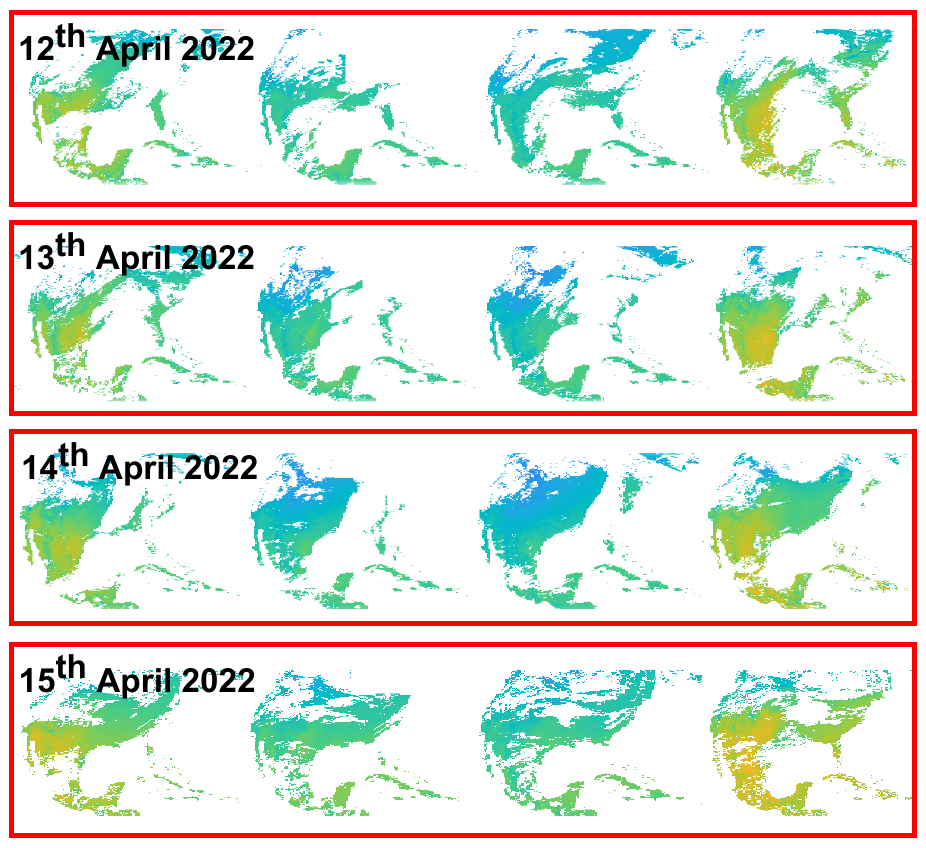}
}
\hfill
\subfigure[UBFC~\citep{bobbia2019unsupervised}]{
    \label{appendix-subfig:ubfc}
    \includegraphics[height=0.41\textwidth]{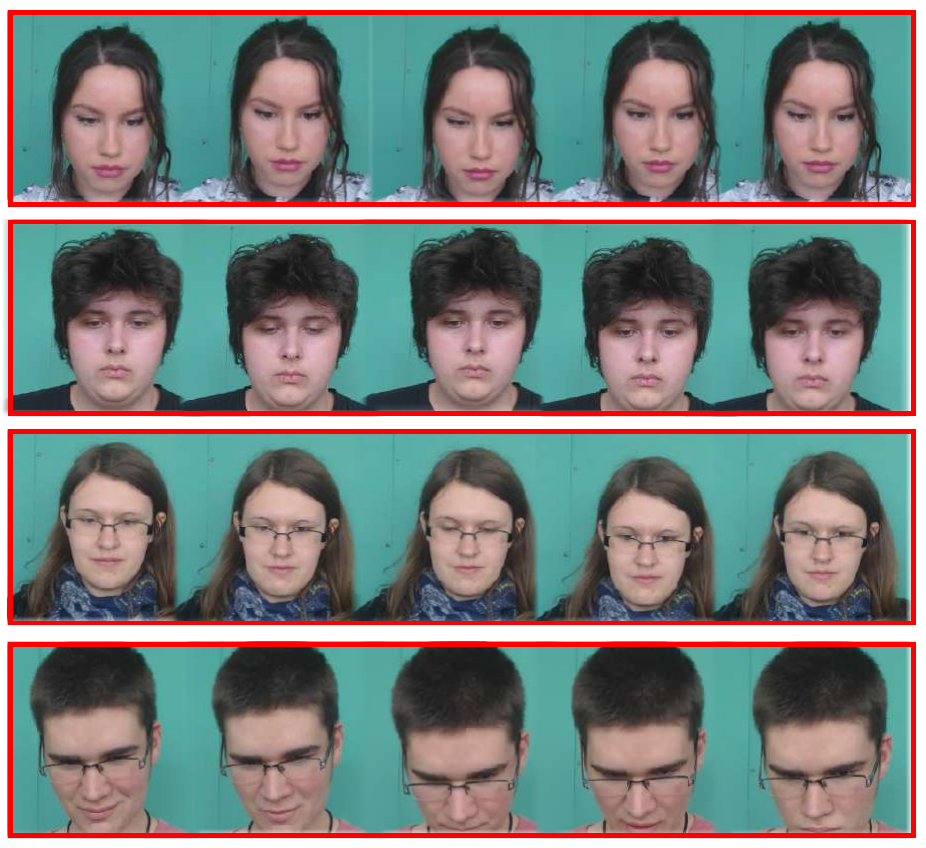}
}
\\
\subfigure[PURE~\citep{stricker2014non}]{
    \label{appendix-subfig:pure}
    \includegraphics[height=0.41\textwidth]{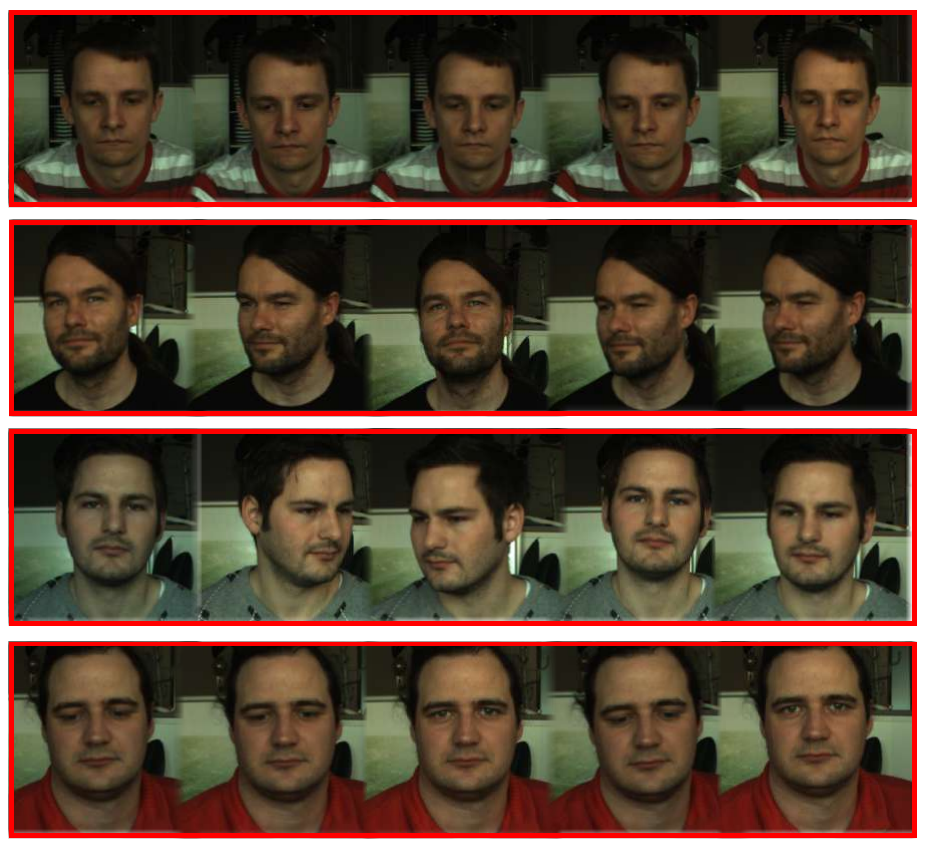}
}
\hfill
\subfigure[Countix~\citep{dwibedi2020counting}]{
    \label{appendix-subfig:countix}
    \includegraphics[height=0.41\textwidth]{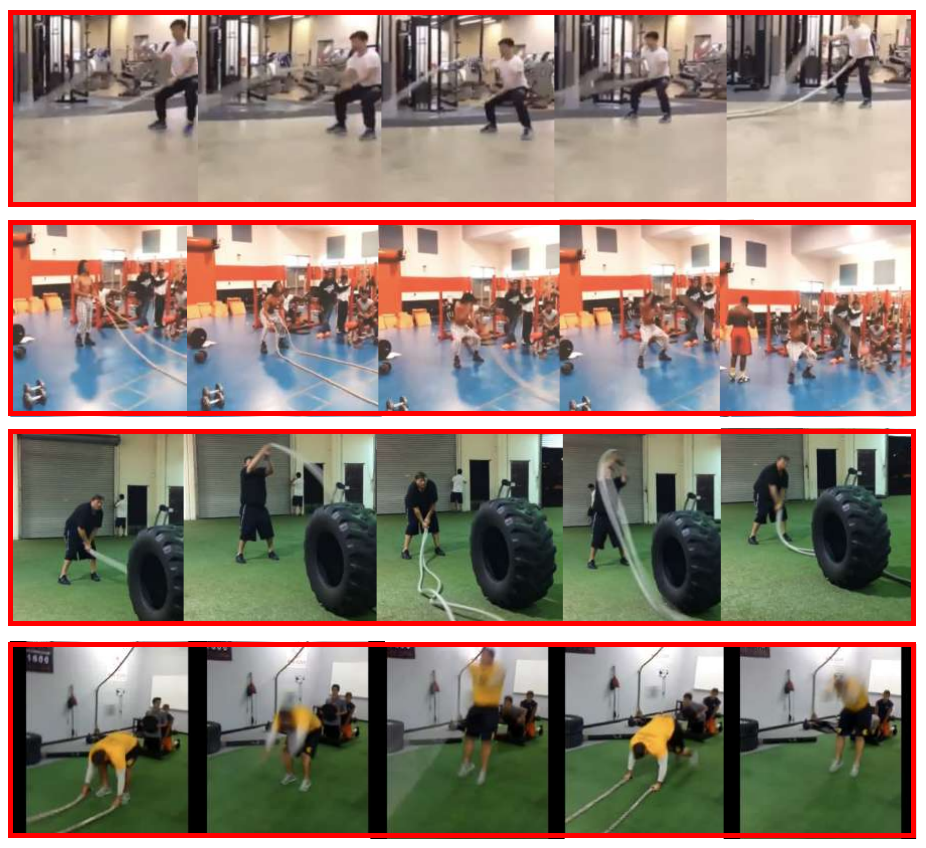}
}
\vspace{-0.3cm}
\caption{\small 
\textbf{Examples of sequences from the datasets used in our experiments.}
}
\label{appendix-fig:dataset-examples}
\vspace{-0.2cm}
\end{figure*}

\section{Dataset Details}
\label{appendix-sec:dataset-details}

In this section, we provide the detailed information of the six datasets we used in our experiments.
Fig.~\ref{appendix-fig:dataset-examples} shows examples of each dataset, and Table \ref{appendix:table:datasets} provides the statistics of each dataset.

\textbf{RotatingDigits} \emph{(Synthetic Dataset)}. We create RotatingDigits, a synthetic periodic learning dataset of rotating MNIST digits~\citep{deng2012mnist}, where samples are created with the original digits rotating on a plain background at rotational frequencies between $0.5$Hz and $5$Hz.
The training set consists of $1,000$ rotating video clips ($100$ samples per digit number), each sample with a frame length of $150$ and a sampling rate of $30$Hz.
The test set consists of $2,000$ rotating video clips ($200$ samples per digit number).

\textbf{SCAMPS} \emph{(Human Physiology)}. The SCAMPS dataset \citep{mcduff2022scamps} contains $2,800$ synthetic videos of avatars with realistic peripheral blood flow and breathing. The faces are synthesized using a blendshape-based rig with $7,667$ vertices and $7,414$ polygons and the identity basis is learned from a set of high-quality facial scans. These texture maps were sampled from 511 facial scans of subjects. The distribution of gender, age and ethnicity of the subjects who provided the facial scans can be found in~\citep{wood2021fake}. Blood flow is simulated by adjusting properties of the physically-based shading material\footnote{\url{https://www.blender.org/}}. We randomly divide the whole dataset into training ($2,000$ samples), validation ($400$ samples), and test ($400$ samples) set. Each video clip has a frame length of $600$ and a sampling rate of $30$Hz.

\textbf{UBFC} \emph{(Human Physiology)}. The UBFC dataset \citep{bobbia2019unsupervised} contains a total of $42$ videos from $42$ subjects. The videos were recorded using a \texttt{Logitech C920 HD Pro} at $30$Hz. A pulse oximeter was used to obtain the gold-standard PPG data ($30$Hz). The raw resolution is $640\times 480$ and videos are recorded in a uncompressed $8$-bit RGB format. We postprocess the videos by cropping the face region and resizing them to $36\times 36$. We manually divide each video into non-overlapping chunks \citep{liu2020multi} with a window size $180$ frames ($6$ seconds). The resulting number of training and test samples are $518$ and $106$, respectively.

\textbf{PURE} \emph{(Human Physiology)}. The PURE dataset \citep{stricker2014non} includes $60$ videos from $10$ subjects ($8$ male, $2$ female). The subjects were asked to seat in front of the camera at an average distance of $1.1$ meters and lit from the front with ambient natural light through a window. Each subject was then instructed to perform six tasks with varying levels of head motion such as slow/fast translation between camera plane and head motion as well as small/medium head rotations. Gold-standard measurements were collected with a pulse oximeter at $60$Hz. The raw video resolution is $640\times 480$. We postprocess the videos by cropping the face region and resizing them to $36\times 36$, and downsample the ground-truth PPG signal to $30$Hz from $60$Hz. We manually divide each video into non-overlapping chunks \citep{liu2020multi} with a window size $180$ frames ($6$ seconds). The resulting number of training and test samples are $1,028$ and $226$, respectively.


\textbf{Countix} \emph{(Action Counting)}. The Countix dataset \citep{dwibedi2020counting} is a subset of the Kinetics~\citep{kay2017kinetics} dataset annotated with segments of repeated actions and corresponding counts. The creators crowdsourced the labels for repetition segments and counts for the selected classes. We further filter out videos that have a frame length shorter than $200$, and make all videos have a fixed length of $200$ frames. The resulting dataset has $1,712$ training samples, $457$ validation samples, and $963$ test samples, with a resolution of $96\times 96$.

\textbf{Land Surface Temperature (LST)} \emph{(Satellite Sensing)}.
Land surface temperature is an indicator of the Earth surface energy budget and is widely required in applications of hydrology, meteorology and climatology. It is of fundamental importance to the net radiation budget at the Earth's surface and for monitoring the state of crops and vegetation, as well as an important indicator of both the greenhouse effect and the energy flux between the atmosphere and earth surface.
We created a snapshot of data from the \texttt{NOAA GOES-16 Level 2 LST product} comprising of hourly land surfaces temperature outputs over the continental United States (CONUS). The LST measurements are sampled hourly over a $100$ day period leading to $2,400$ LST maps at a resolution of $1,500 \times 2,500$. As the spatial resolution is high, we divide each map into four quarters (\texttt{North-West US}, \texttt{North-East US}, \texttt{South-West US}, and \texttt{South-East US}).
We create each input sample using a window size of $100$ frames with a step size of $24$ (a day). The target signal is the temperature time series of the future $100$ frames.
The resulting dataset has $276$ training samples and $92$ test samples, with a spatial resolution of $100\times 100$.

\begin{table}[!t]
\setlength{\tabcolsep}{5pt}
\caption{\small
\textbf{Detailed statistics of the datasets used in our experiments.}}
\vspace{-1.5pt}
\label{appendix:table:datasets}
\small
\begin{center}
\adjustbox{max width=\textwidth}{
\begin{tabular}{llccccc}
\toprule[1.5pt]
& Targets & Sampling freq. & Frame length & \# Training set &  \# Val. set & \# Test set  \\
\midrule \midrule
RotatingDigits & Rotation frequency & 30Hz & 150 & 1,000 & $-$ & 2,000 \\\midrule
SCAMPS~\citep{mcduff2022scamps} & Heart rate & 30Hz & 600 & 2,000 & 400 & 400 \\\midrule
UBFC~\citep{bobbia2019unsupervised} & Heart rate & 30Hz & 180 & 518 & $-$ & 106 \\\midrule
PURE~\citep{stricker2014non} & Heart rate & 30Hz & 180 & 1,028 & $-$ & 266 \\\midrule
Countix~\citep{dwibedi2020counting} & Action counts & 20$\sim$30Hz & 200 & 1,712 & 457 & 963 \\\midrule
LST & Temperature & Hourly & 100 & 276 & $-$ & 92 \\
\bottomrule[1.5pt]
\end{tabular}}
\end{center}
\end{table}

\section{Experimental Settings}
\label{appendix-sec:exp-settings}

\subsection{Competing Algorithms}
\label{appendix-subsec:baselines}

We employ the following state-of-the-art SSL algorithms for comparisons.

\textbf{SimCLR} \citep{chen2020simple}.
SimCLR learns feature representations by contrasting images with data augmentation. The positive pairs are constructed by sampling two images with different augmentations on one instance. The negative pairs are sampled from two different images.

\textbf{MoCo v2} \citep{he2020momentum}.
MoCo learns feature representations by building large dictionaries along with a contrastive loss. MoCo maintains the dictionary as a queue of data samples by enqueuing the current mini-batch and dequeuing the oldest mini-batch. The keys are encoded by a slowly progressing encoder with a momentum moving average and the query encoder.

\textbf{BYOL} \citep{grill2020bootstrap}.
BYOL leverages two neural networks to learn the feature representations: the online and target networks. The online network has an encoder, a projector, and a predictor while the target network shares the same architecture but with a different set of weights. The online network is trained by the regression targets provided by the target network.

\textbf{CVRL} \citep{qian2021spatiotemporal}.
Contrastive Video Representation Learning (CVRL) is a self-supervised learning framework that learns spatial-temporal features representations from unlabelled videos. CVRL generates positive pairs by adding temporally consistent spatial augmentation on one videos clip and generate negative pairs by sampling two different video clips. The goal of contrastive loss is to minimize the embedding distance from the positive augmented video clips but maximize the distance from negative video clips.

\subsection{Implementation Details}
\label{appendix-subsec:implementation-details}

We describe the implementation details in this section. We first introduce parameters that are fixed to be the same across all methods, then detail the specific parameters for each dataset.

For all SSL methods, we follow the literature \citep{chen2020simple,he2020momentum} and apply the same standard data augmentations in contrastive learning. For spatial augmentations, we employ \texttt{random\_crop\_resize}, \texttt{random\_brightness}, \texttt{random\_gaussian\_blur}, \texttt{random\_grayscale}, and \texttt{random\_flip\_left\_right}.
For temporal augmentations, we mainly employ \texttt{random\_reverse} and \texttt{random\_delay} (with shorter clip subsampling \citep{qian2021spatiotemporal}). Unless specified, all augmentation hyper-parameters follow the original setup of each method.

\textbf{RotatingDigits.}
On RotatingDigits, we adopt the network architecture as a simple 3D variant of the MNIST CNN used in \citep{yang2022multi,gulrajani2020domainbed}.
In the supervised setting, we train all models for $20$ epochs using the Adam optimizer~\citep{kingma2014adam}, with an initial learning rate of $10^{-3}$ and then decayed by $0.1$ at the 12-th and 16-th epoch, respectively.
We fix the batch size as $64$ and use the checkpoint at the last epoch as the final model for evaluation.
In the self-supervised setting, we train all models for $60$ epochs, which ensures convergence for all tested algorithms. We again employ the Adam optimizer and decay the learning rating at the 40-th and 50-th epoch, respectively. Other training hyper-parameters remain unchanged.

\textbf{SCAMPS.}
Similar to RotatingDigits, we employ the same 3D CNN architecture for all the SCAMPS experiments.
In the supervised setting, we train all of the models for $30$ epochs using the Adam optimizer, with an initial learning rate of $10^{-3}$ and then decayed by $0.1$ at the 20-th and 25-th epoch, respectively.
We fix the batch size as $32$ and use the last checkpoint for final evaluation.
In the self-supervised setting, we follow the same training regime in RotatingDigits as described in the previous section.

\textbf{UBFC \& PURE.}
Following~\citep{liu2020multi, liu2021metaphys}, we use the temporal shift convolution attention network (TS-CAN) as our backbone model. To adapt TS-CAN on \simper, we remove the attention branch and make a variant of TS-CAN which only requires $3$-channel as the input instead of $6$-channel. In the supervised setting, we use the Adam optimizer, learning rate of $10^{-3}$ and train the network for a total of $10$ epochs. On the inner-dataset evaluation (i.e., test and validation are the same), we use last epoch from the training of $80\%$ for the dataset and evaluate the pre-trained model on the last $20\%$ dataset. On the cross-dataset evaluation, we use $80\%$ of the dataset for training and $20\%$ for checkpoint selection then evaluate the pre-trained model on a different dataset.
In the self-supervised setting, all other parameters remain unchanged except that we train for $60$ epochs to ensure the SSL loss converges for all algorithms.

\textbf{Countix.}
We use a ResNet-3D-18 \citep{tran2018resnet3d,he2016deep} architecture for all Countix experiments, which is widely used for video-based vision tasks.
In the supervised setting, we train all models for $90$ epochs using the Adam optimizer with an initial learning rate of $10^{-3}$ and then decayed by $0.1$ at the 60-th and 80-th epoch. We fix the batch size as $32$ for all experiments.
In the self-supervised setting, we train all models for $200$ epochs, and leave other parameters unchanged.

\textbf{LST.}
Similar to Countix, we use the ResNet-3D-18 \citep{tran2018resnet3d} network architecture for LST experiments.
In the supervised setting, we train all models for $30$ epochs using the Adam optimizer with a learning rate of $10^{-3}$ and a batch size of $16$.
In the self-supervised setting, we train all models for $60$ epochs while having other hyper-parameters the same for all methods.

\subsection{Evaluation Metrics}
\label{appendix-subsec:eval-metrics}

We describe in detail all the evaluation metrics we used in our experiments.

\textbf{MAE.} The mean absolute error~(MAE) is defined as $\frac{1}{N}\sum_{i=1}^{N}|y_i-\hat{y}_i|$, which represents the averaged absolute difference between the ground truth and predicted values over all samples.

\textbf{MAPE.} The mean absolute percentage error~(MAPE) is defined as $\frac{1}{N}\sum_{i=1}^{N}|\frac{y_i-\hat{y}_i}{y_i}|$, which assesses the averaged relative differences between the ground truth and predicted values over all samples. 

\textbf{GM.} We use error Geometric Mean (\textbf{GM}) as another evaluation metric \citep{yang2021delving}. GM is defined as $(\prod_{i=1}^N e_i)^{\frac{1}{N}}$, where $e_i\triangleq|y_i-\hat{y}_i|$ represents the $L_1$ error of each sample. GM aims to characterize the fairness (uniformity) of model predictions using the geometric mean instead of the arithmetic mean over the prediction errors.

\textbf{Pearson correlation $\rho$.} We employ Pearson correlation for performance evaluation on LST, where Pearson correlation evaluates the linear relationship between predictions and corresponding ground truth values.

\section{Additional Results and Analysis}
\label{appendix-sec:additional-results}

\subsection{Data Efficiency w.r.t. Reduced Training Data}
\label{appendix-subsec:data-efficiency-reduced-training-data}

We provide quantitative results to verify the data efficiency of \simper in the presence of reduced training data. Specifically, we use SCAMPS dataset, and vary the training dataset size from $100\%$ to only $5\%$, and use it for both pre-training and fine-tuning.
We show the final performance in Table \ref{appendix:table:reduced-training-data}, where \simper is able to achieve consistent performance gains compared to baselines when the dataset size varies.
Furthermore, the gains are more significant when the dataset size is smaller (e.g., $5\%$), demonstrating that \simper is particularly robust to reduced training data.

\vspace{0.5cm}
\begin{table}[h]
\setlength{\tabcolsep}{3pt}
\caption{\small 
\textbf{Data efficiency w.r.t. reduced training data.} We vary the training dataset size of SCAMPS (size fixed for both pre-training and fine-tuning), and show the final fine-tuning performance of different methods.
}
\vspace{-5pt}
\label{appendix:table:reduced-training-data}
\small
\begin{center}
\adjustbox{max width=\textwidth}{
\begin{tabular}{lcccccccccc}
\toprule[1.5pt]
Dataset size & \multicolumn{2}{c}{100\%} & \multicolumn{2}{c}{50\%} & \multicolumn{2}{c}{20\%} & \multicolumn{2}{c}{10\%} & \multicolumn{2}{c}{5\%} \\
\cmidrule(lr){2-3} \cmidrule(lr){4-5} \cmidrule(lr){6-7} \cmidrule(lr){8-9} \cmidrule(lr){10-11}
Metrics & MAE$^\downarrow$ & MAPE$^\downarrow$ & MAE$^\downarrow$ & MAPE$^\downarrow$ & MAE$^\downarrow$ & MAPE$^\downarrow$ & MAE$^\downarrow$ & MAPE$^\downarrow$ & MAE$^\downarrow$ & MAPE$^\downarrow$ \\ \midrule\midrule
\textsc{Supervised} & 3.61 & 5.33 & 3.85 & 5.60 & 4.57 & 7.16 & 7.13 & 10.08 & 12.24 & 15.42 \\ \midrule
\textsc{SimCLR}~\citep{chen2020simple} & 4.96 & 6.92 & 6.55 & 9.39 & 6.01 & 9.25 & 7.63 & 10.19 & 13.75 & 15.72 \\[1.2pt]
\textsc{CVRL}~\citep{qian2021spatiotemporal} & 5.52 & 7.34 & 3.66 & 5.64 & 4.86 & 7.77 & 7.08 & 9.45 & 14.11 & 15.91 \\[1.2pt]
\grayrow
\textbf{\textsc{SimPer}} & \textbf{3.27} & \textbf{4.89} & \textbf{3.38} & \textbf{5.24} & \textbf{3.93} & \textbf{5.67} & \textbf{4.65} & \textbf{7.06} & \textbf{4.75} & \textbf{7.64} \\
\midrule
\textsc{Gains vs. Supervised} & \textcolor{darkgreen}{\texttt{+}\textbf{0.34}} & \textcolor{darkgreen}{\texttt{+}\textbf{0.44}} & \textcolor{darkgreen}{\texttt{+}\textbf{0.47}} & \textcolor{darkgreen}{\texttt{+}\textbf{0.36}} & \textcolor{darkgreen}{\texttt{+}\textbf{0.64}} & \textcolor{darkgreen}{\texttt{+}\textbf{1.49}} & \textcolor{darkgreen}{\texttt{+}\textbf{2.48}} & \textcolor{darkgreen}{\texttt{+}\textbf{3.02}} & \textcolor{darkgreen}{\texttt{+}\textbf{7.49}} & \textcolor{darkgreen}{\texttt{+}\textbf{7.78}} \\
\bottomrule[1.5pt]
\end{tabular}}
\end{center}
\end{table}
\vspace{0.3cm}

\subsection{Amount of Labeled Data for Fine-tuning}
\label{appendix-subsec:data-efficiency-amount-labeled-data}

We investigate the impact of the amount of labeled data for fine-tuning. Specifically, we use the whole training set of SCAMPS as the unlabeled dataset, and vary the labeled data fraction for fine-tuning.
As Table \ref{appendix:table:amount-labeled-data} confirms, when the amount of labeled data is limited for fine-tuning, \simper still substantially outperforms baselines by a large margin, achieving a $67\%$ relative improvement in MAE even when the labeled data fraction is only $5\%$.
The results again demonstrate that \simper is data efficient in terms of the amount of labeled data available.

\vspace{0.5cm}
\begin{table}[h]
\setlength{\tabcolsep}{3pt}
\caption{\small 
\textbf{Data efficiency w.r.t. amount of labeled data for fine-tuning.} We use all data from SCAMPS as unlabeled training set for self-supervised pre-training, and vary size of labeled data for fine-tuning.
}
\vspace{-5pt}
\label{appendix:table:amount-labeled-data}
\small
\begin{center}
\adjustbox{max width=\textwidth}{
\begin{tabular}{lcccccccccc}
\toprule[1.5pt]
Labeled data fraction & \multicolumn{2}{c}{100\%} & \multicolumn{2}{c}{50\%} & \multicolumn{2}{c}{20\%} & \multicolumn{2}{c}{10\%} & \multicolumn{2}{c}{5\%} \\
\cmidrule(lr){2-3} \cmidrule(lr){4-5} \cmidrule(lr){6-7} \cmidrule(lr){8-9} \cmidrule(lr){10-11}
Metrics & MAE$^\downarrow$ & MAPE$^\downarrow$ & MAE$^\downarrow$ & MAPE$^\downarrow$ & MAE$^\downarrow$ & MAPE$^\downarrow$ & MAE$^\downarrow$ & MAPE$^\downarrow$ & MAE$^\downarrow$ & MAPE$^\downarrow$ \\ \midrule\midrule
\textsc{Supervised} & 3.61 & 5.33 & 3.85 & 5.60 & 4.57 & 7.16 & 7.13 & 10.08 & 12.24 & 15.42 \\ \midrule
\textsc{SimCLR}~\citep{chen2020simple} & 4.96 & 6.92 & 4.92 & 7.09 & 5.57 & 8.46 & 7.82 & 10.53 & 13.21 & 15.64 \\[1.2pt]
\textsc{CVRL}~\citep{qian2021spatiotemporal} & 5.52 & 7.34 & 3.79 & 5.83 & 4.83 & 7.71 & 6.82 & 9.06 & 12.18 & 13.25 \\[1.2pt]
\grayrow
\textbf{\textsc{SimPer}} & \textbf{3.27} & \textbf{4.89} & \textbf{3.32} & \textbf{5.13} & \textbf{3.58} & \textbf{5.44} & \textbf{3.98} & \textbf{5.81} & \textbf{4.02} & \textbf{6.27} \\
\midrule
\textsc{Gains vs. Supervised} & \textcolor{darkgreen}{\texttt{+}\textbf{0.34}} & \textcolor{darkgreen}{\texttt{+}\textbf{0.44}} & \textcolor{darkgreen}{\texttt{+}\textbf{0.53}} & \textcolor{darkgreen}{\texttt{+}\textbf{0.47}} & \textcolor{darkgreen}{\texttt{+}\textbf{0.99}} & \textcolor{darkgreen}{\texttt{+}\textbf{1.72}} & \textcolor{darkgreen}{\texttt{+}\textbf{3.15}} & \textcolor{darkgreen}{\texttt{+}\textbf{4.27}} & \textcolor{darkgreen}{\texttt{+}\textbf{8.22}} & \textcolor{darkgreen}{\texttt{+}\textbf{9.15}} \\
\bottomrule[1.5pt]
\end{tabular}}
\end{center}
\end{table}
\vspace{0.3cm}

\subsection{Robustness to Spurious Correlations}
\label{appendix-subsec:spurious}

We provide detailed quantitative results for the spurious correlations experiment in Section \ref{subsec:exp-spurious-corr}.
Recall that SimCLR is easy to learn information that is spuriously correlated in the training data, and the learned representations do not generalize. Table \ref{appendix:table:spurious} further confirms the observation, where SimCLR achieves bad feature evaluation results with large MAE \& MAPE errors.

In contrast, \simper is able to learn the underlying frequency information even in the presence of strong spurious correlations, obtaining substantially smaller errors compared to SimCLR.
The results demonstrate that \simper is robust to spurious correlations, and can learn robust representations that generalize.

\begin{table}[!t]
\setlength{\tabcolsep}{10pt}
\caption{\small
\textbf{Feature evaluation results on RotatingDigits with spurious correlations in training data.} Quantitative results in addition to Fig.~\ref{fig:spurious} further verify that state-of-the-art SSL methods (e.g., SimCLR) are vulnerable to spurious correlations, and could easily learn information that is irrelevant to periodicity; In contrast, \simper learns desirable periodic representations that are robust to spurious correlations.}
\vspace{-1pt}
\label{appendix:table:spurious}
\small
\begin{center}
\adjustbox{max width=0.7\textwidth}{
\begin{tabular}{lcccc}
\toprule[1.5pt]
 & \multicolumn{2}{c}{FFT} & \multicolumn{2}{c}{1-NN} \\
\cmidrule(lr){2-3} \cmidrule(lr){4-5}
Metrics & MAE$^\downarrow$ & MAPE$^\downarrow$ & MAE$^\downarrow$ & MAPE$^\downarrow$ \\ \midrule\midrule
\textsc{SimCLR}~\citep{chen2020simple} & 3.06 & 125.48 & 1.49 & 80.28 \\[1.2pt]
\grayrow
\textbf{\textsc{SimPer}} & \textbf{0.36} & \textbf{15.04} & \textbf{0.78} & \textbf{27.03} \\
\midrule
\textsc{Gains} & \textcolor{darkgreen}{\texttt{+}\textbf{2.70}} & \textcolor{darkgreen}{\texttt{+}\textbf{110.44}} & \textcolor{darkgreen}{\texttt{+}\textbf{0.71}} & \textcolor{darkgreen}{\texttt{+}\textbf{53.25}} \\
\bottomrule[1.5pt]
\end{tabular}}
\end{center}
\end{table}

\subsection{Ablation Studies for SimPer}
\label{appendix-subsec:ablation}

In this section, we perform extensive ablation studies on \simper to investigate the effect of different design choices as well as its hyper-parameter stability.

\vspace{-0.1cm}
\subsubsection{Range of Periodicity-Variant Frequency Augmentation}
\label{appendix-subsubsec:ablation-freq-range}

We study the effect of using different ranges of the variant speed augmentations in \simper. We use the SCAMPS dataset, and vary the speed range during \simper pre-training.
As Table \ref{appendix:table:ablation-freq-range} reports, using different speed ranges does not change the downstream performance by much, where all the results outperform the supervised baseline by a notable margin.

\vspace{0.4cm}
\begin{table}[ht]
\setlength{\tabcolsep}{7pt}
\caption{\small
\textbf{Ablation study on the range of speed (frequency) augmentation.} Default settings used in the main experiments for \simper are marked in \colorbox{baselinecolor}{gray}.}
\vspace{-2pt}
\label{appendix:table:ablation-freq-range}
\small
\begin{center}
\adjustbox{max width=0.9\textwidth}{
\begin{tabular}{l|cccc|c}
\toprule[1.5pt]
\textsc{Speed Range} & $[0.5, 1.5]$ & $[0.8, 1.8]$ & \graycell{$[0.5, 2]$} & $[0.5, 3]$ & \gc{Supervised} \\
\midrule
MAPE$^\downarrow$ & 4.97 & 4.92 & \graycell{4.89} & 4.98 & \gc{5.33} \\
\bottomrule[1.5pt]
\end{tabular}}
\end{center}
\end{table}
\vspace{0.2cm}

\subsubsection{Number of Periodicity-Variant Augmented Views}
\label{appendix-subsubsec:ablation-num-augs}

We study the effect of different number of periodicity-variant augmented views $M$ on \simper. We again employ the SCAMPS dataset, and vary the number of augmented views as $M\in \{3,5,10,20\}$.
Table \ref{appendix:table:ablation-num-views} shows the results, where we can observe a clear trend of decreased error rates when increasing $M$.
Yet, when $M\geq 5$, the benefits of increasing $M$ gradually diminish, indicating that a moderate $M$ might be enough for the task. In the experiments of all tested datasets, to balance the efficiency while maintaining the contrastive ability, we set $M=10$ by default.

\vspace{0.5cm}
\begin{table}[ht]
\setlength{\tabcolsep}{7pt}
\caption{\small
\textbf{Ablation study on the number of periodicity-variant augmented views.} Default settings used in the main experiments for \simper are marked in \colorbox{baselinecolor}{gray}.}
\label{appendix:table:ablation-num-views}
\small
\begin{center}
\adjustbox{max width=0.9\textwidth}{
\begin{tabular}{l|cccc|c}
\toprule[1.5pt]
\textsc{Num. Views} & 3 & 5 & \graycell{10} & 20 & \gc{Supervised} \\
\midrule
MAPE$^\downarrow$ & 5.12 & 4.96 & \graycell{4.89} & 4.87 & \gc{5.33} \\
\bottomrule[1.5pt]
\end{tabular}}
\end{center}
\end{table}
\vspace{0.2cm}

\subsubsection{Choices of Different Similarity Metrics}
\label{appendix-subsubsec:similarity-metrics}

We investigate the impact of different choices of periodic similarity measures introduced in Section \ref{subsec:period-feat-sim}.
Specifically, we study three concrete instantiations of periodic similarity measures: \textbf{MXCorr}, \textbf{nPSD ($\cos(\cdot)$)}, and \textbf{nPSD ($L_2$)}.
As Table \ref{appendix:table:ablation-similarity-metrics} shows, \simper is robust to all aforementioned periodic similarity measures, achieving similar downstream performances.
The results also demonstrate the effectiveness of the proposed similarity measures in periodic learning.

\begin{table}[!t]
\setlength{\tabcolsep}{7pt}
\caption{\small
\textbf{Ablation study on the choices of different periodic similarity measures.} Default settings used in the main experiments for \simper are marked in \colorbox{baselinecolor}{gray}.}
\vspace{-2pt}
\label{appendix:table:ablation-similarity-metrics}
\small
\begin{center}
\adjustbox{max width=0.95\textwidth}{
\begin{tabular}{l|ccc|c}
\toprule[1.5pt]
\textsc{Similarity Metrics} & \graycell{MXCorr} & nPSD ($\cos(\cdot)$) & nPSD ($L_2$) & \gc{Supervised} \\
\midrule
MAPE$^\downarrow$ & \graycell{4.89} & 4.88 & 4.92 & \gc{5.33} \\
\bottomrule[1.5pt]
\end{tabular}}
\end{center}
\end{table}

\subsubsection{Effectiveness of the Generalized Contrastive Loss}
\label{appendix-subsubsec:gen-con-loss}

We assess the effectiveness of the generalized contrastive loss, as compared to the classic InfoNCE contrastive loss.
Table \ref{appendix:table:ablation-gen-con-loss} highlights the results over all six datasets, where consistent gains can be obtained when using the generalized contrastive loss in \simper formulation.

\vspace{0.4cm}
\begin{table}[ht]
\setlength{\tabcolsep}{4pt}
\caption{\small
\textbf{Ablation study on the effectiveness of using generalized contrastive loss in \simper.} We show the feature evaluation results (FFT, MAE$^\downarrow$) with and without generalized contrastive loss across different datasets. Note that generalized contrastive loss with no continuity considered degenerates to InfoNCE~\citep{oord2018representation}.}
\label{appendix:table:ablation-gen-con-loss}
\small
\begin{center}
\adjustbox{max width=0.95\textwidth}{
\begin{tabular}{lcccccc}
\toprule[1.5pt]
& RotatingDigits & SCAMPS & UBFC & PURE & Countix & LST \\
\midrule \midrule
\simper (InfoNCE) & \textbf{0.23} & 18.27 & 9.53 & 15.74 & 2.42 & \textbf{4.84} \\[1.2pt]
\grayrow
\simper (Generalized) & \textbf{0.22} & \textbf{14.45} & \textbf{8.78} & \textbf{13.97} & \textbf{2.06} & \textbf{4.84} \\
\midrule
Gains & \textcolor{darkgreen}{\texttt{+}\textbf{0.01}} & \textcolor{darkgreen}{\texttt{+}\textbf{3.82}} & \textcolor{darkgreen}{\texttt{+}\textbf{0.75}} & \textcolor{darkgreen}{\texttt{+}\textbf{1.77}} & \textcolor{darkgreen}{\texttt{+}\textbf{0.36}} & \textcolor{darkgreen}{\texttt{+}\textbf{0.00}} \\
\bottomrule[1.5pt]
\end{tabular}}
\end{center}
\end{table}
\vspace{0.5cm}

\subsubsection{Choices of Different Input Sequence Lengths}
\label{appendix-subsubsec:frame-lengths}

Finally, we investigate the effect of different sequence lengths on the final performance in periodic learning. To make the observations more general and comprehensive, we choose three datasets from different domains (i.e., RotatingDigits, SCAMPS, and LST) to study the effect of sequence length. We fix all the experimental setups the same as in Appendix \ref{appendix-sec:dataset-details} \& \ref{appendix-sec:exp-settings}, and only vary the frame/sequence lengths with different yet reasonable choices for each dataset.

As highlighted from Table \ref{appendix:table:ablation-frame-lengths}, the results illustrate the following interesting observations:
\begin{Itemize}
    \item For ``clean'' periodic learning datasets with the periodic targets being the only dominating signal (i.e., RotatingDigits), using different frame lengths do not inherently change the final result.
    \vspace{1em}
    \item For dataset with relatively high SNR (i.e., LST), \simper is also robust to different frame lengths. The supervised results however are worse with shorter clips, which could be attributed to the fact that less information is used in the input.
    \vspace{1em}
    \item Interestingly, for datasets where other periodic signals might exist (i.e., SCAMPS), using shorter (but with reasonable length) videos seems to slightly improve the performance of \simper. We hypothesize that for a complex task such as video-based human physiological measurement, some videos may contain multiple periodic processes (e.g., PPG, breathing, blinking, etc.). A smaller frame length may not be enough to capture some of the ``slow'' periodic processes (e.g., breathing), thus the features learned by \simper can become even more representative for PPG or heart beats estimation. Nevertheless, the differences between various choices are still small, indicating that \simper is pretty robust to different frame lengths.
\end{Itemize}

\begin{table}[!t]
\setlength{\tabcolsep}{8pt}
\caption{\small
\textbf{Ablation study on the input sequence lengths.} We show the fine-tune evaluation results (MAE$^\downarrow$) using different yet reasonable sequence lengths across various datasets.}
\label{appendix:table:ablation-frame-lengths}
\small
\begin{center}
\adjustbox{max width=0.95\textwidth}{
\begin{tabular}{lccccccccc}
\toprule[1.5pt]
           & \multicolumn{3}{c}{RotatingDigits} & \multicolumn{3}{c}{SCAMPS} & \multicolumn{3}{c}{LST} \\
\cmidrule(lr){2-4} \cmidrule(lr){5-7} \cmidrule(lr){8-10}
\# Frames  & 150        & 120       & 90        & 600     & 450     & 300    & 100    & 80     & 60    \\ \midrule\midrule
\textsc{Supervised} & 0.72       & 0.71      & 0.72      & 3.61    & 3.57    & 3.63   & 1.54   & 1.56   & 1.61  \\[1.2pt]
\grayrow
\textsc{\textbf{SimPer}}     & \textbf{0.20}       & \textbf{0.19}      & \textbf{0.20}      & \textbf{3.27}    & \textbf{3.11}    & \textbf{3.12}   & \textbf{1.47}   & \textbf{1.47}   & \textbf{1.48}  \\
\midrule
\textsc{Gains} & \textcolor{darkgreen}{\texttt{+}\textbf{0.52}} & \textcolor{darkgreen}{\texttt{+}\textbf{0.52}} & \textcolor{darkgreen}{\texttt{+}\textbf{0.52}} & \textcolor{darkgreen}{\texttt{+}\textbf{0.34}} & \textcolor{darkgreen}{\texttt{+}\textbf{0.46}} & \textcolor{darkgreen}{\texttt{+}\textbf{0.51}} & \textcolor{darkgreen}{\texttt{+}\textbf{0.07}} & \textcolor{darkgreen}{\texttt{+}\textbf{0.09}} & \textcolor{darkgreen}{\texttt{+}\textbf{0.13}} \\
\bottomrule[1.5pt]
\end{tabular}}
\end{center}
\end{table}
\vspace{0.2cm}

\subsection{Comparisons and Compatibility with SOTA Supervised Learning Methods}
\label{appendix-subsec:sota-sup-algos}

As motivated in the main paper, for each specific periodic learning application, supervised learning methods \citep{dwibedi2020counting,mcduff2022scamps,liu2023efficientphys} have achieved remarkably good results via incorporating certain domain knowledge tailored for a specific task.
Therefore, we provide additional results and comparisons using SOTA algorithms on each of the tested dataset. 
In the following, we show existing SOTA baselines and demonstrate that \simper could further boost the performance when jointly applied.

\vspace{0.4cm}
\begin{table}[ht]
\setlength{\tabcolsep}{10pt}
\caption{\small
\textbf{Compatibility of \simper with SOTA supervised techniques across different datasets.} SOTA refers to RepNet \citep{dwibedi2020counting} on Countix, and refers to EfficientPhys \citep{liu2023efficientphys} on SCAMPS, UBFC \& PURE. \simper delivers robust performance and complements the performance of SOTA models.}
\vspace{-1pt}
\label{appendix:table:sota-sup-algos}
\small
\begin{center}
\adjustbox{max width=0.98\textwidth}{
\begin{tabular}{lcccccccc}
\toprule[1.5pt]
 & \multicolumn{2}{c}{Countix} & \multicolumn{2}{c}{SCAMPS} & \multicolumn{2}{c}{UBFC} & \multicolumn{2}{c}{PURE} \\
\cmidrule(lr){2-3} \cmidrule(lr){4-5} \cmidrule(lr){6-7} \cmidrule(lr){8-9}
Metrics & MAE$^\downarrow$ & GM$^\downarrow$ & MAE$^\downarrow$ & MAPE$^\downarrow$ & MAE$^\downarrow$ & MAPE$^\downarrow$ & MAE$^\downarrow$ & MAPE$^\downarrow$ \\ \midrule\midrule
\textsc{SOTA}  & 1.03 & 0.41 & 2.42 & 4.10 & 4.14 & 3.79 & 2.87 & 2.89 \\ \midrule
\textsc{SimCLR + SOTA} & 1.06 & 0.43 & 2.56  & 4.17 & 4.31 & 4.02 & 2.94 & 3.25 \\[1.2pt]
\grayrow
\textbf{\textsc{SimPer + SOTA}} & \textbf{0.72} & \textbf{0.22} & \textbf{1.96} & \textbf{3.45} & \textbf{3.27} & \textbf{3.06} & \textbf{2.29} & \textbf{2.21} \\
\midrule
\textsc{Gains} & \textcolor{darkgreen}{\texttt{+}\textbf{0.29}} & \textcolor{darkgreen}{\texttt{+}\textbf{0.19}} & \textcolor{darkgreen}{\texttt{+}\textbf{0.46}} & \textcolor{darkgreen}{\texttt{+}\textbf{0.65}} & \textcolor{darkgreen}{\texttt{+}\textbf{0.87}} & \textcolor{darkgreen}{\texttt{+}\textbf{0.73}} & \textcolor{darkgreen}{\texttt{+}\textbf{0.58}} & \textcolor{darkgreen}{\texttt{+}\textbf{0.68}} \\
\bottomrule[1.5pt]
\end{tabular}}
\end{center}
\end{table}
\vspace{0.5cm}

\textbf{Countix.}
In the video repetition counting domain, RepNet \citep{dwibedi2020counting}, a novel neural network architecture that composed of a ResNet-50 encoder and a Transformer based predictor, is proposed to achieve advanced results for repetitious counting in the wild. We verify the compatibility of \simper with RepNet by changing the encoder on Countix to RepNet, and compare with the vanilla supervised training as well as SimCLR.
To ensure a fair and comparable setting, we train RepNet from scratch instead of using ImageNet pre-trained ResNet-50 backbones as in the original paper \citep{dwibedi2020counting}.

\textbf{SCAMPS, UBFC \& PURE.}
In video-based human physiological sensing domain (i.e., SCAMPS, UBFC, and PURE), the main advances in the field have stemmed from better backbone architectures and network components \citep{liu2020multi,liu2023efficientphys,gideon2021way}. In the main paper, for SCAMPS, since it is a synthetic dataset, we employed a simple 3D ConvNet; as for real datasets UBFC and PURE, we used a more advanced backbone model \citep{liu2020multi}. To further demonstrate that \simper can improve upon SOTA methods, we employ a recent architecture, called EfficientPhys \citep{liu2023efficientphys}, which is specialized for learning physiology from videos.

As confirmed in Table \ref{appendix:table:sota-sup-algos}, when jointly applied with SOTA models, \simper can further boost the performance and consistently achieves the best results regardless of datasets and tasks. In contrast, SimCLR is not able to improve upon SOTA supervised learning techniques. The results indicate that \simper is orthogonal to SOTA models for learning periodic targets.

\begin{table}[ht]
\setlength{\tabcolsep}{10pt}
\caption{\small
\textbf{Comparisons between \simper and additional SSL baselines on human physiological measurement datasets.} Compared to customized SSL algorithms in the specific domain, \simper still delivers robust performance and consistently achieves the best results.}
\vspace{-1pt}
\label{appendix:table:sota-ssl-algos-rppg-datasets}
\small
\begin{center}
\adjustbox{max width=0.9\textwidth}{
\begin{tabular}{lcccccc}
\toprule[1.5pt]
 & \multicolumn{2}{c}{SCAMPS} & \multicolumn{2}{c}{UBFC} & \multicolumn{2}{c}{PURE} \\
\cmidrule(lr){2-3} \cmidrule(lr){4-5} \cmidrule(lr){6-7}
Metrics & MAE$^\downarrow$ & MAPE$^\downarrow$ & MAE$^\downarrow$ & MAPE$^\downarrow$ & MAE$^\downarrow$ & MAPE$^\downarrow$ \\ \midrule\midrule
\multicolumn{7}{l}{\emph{\textbf{Without face saliency module:}}} \\ \midrule
\citep{gideon2021way} & 3.53 & 5.26 & 4.98 & 4.61 & 4.18 & 4.70 \\[1.2pt]
\citep{wang2022self} & 3.71 & 5.54 & 5.07 & 4.88 & 4.32 & 4.95 \\[1.2pt]
\grayrow
\textbf{\textsc{SimPer}} & \textbf{3.27} & \textbf{4.89} & \textbf{4.24} & \textbf{3.97} & \textbf{3.89} & \textbf{4.01} \\ \midrule\midrule
\multicolumn{7}{l}{\emph{\textbf{With face saliency module:}}} \\ \midrule
\citep{gideon2021way} & 3.51 & 5.15 & 4.88 & 4.29 & 4.03 & 4.28 \\[1.2pt]
\citep{wang2022self} & 3.61 & 5.40 & 5.02 & 4.86 & 4.07 & 4.33 \\[1.2pt]
\grayrow
\textbf{\textsc{SimPer}} & \textbf{2.94} & \textbf{4.35} & \textbf{4.01} & \textbf{3.68} & \textbf{3.47} & \textbf{3.76} \\
\bottomrule[1.5pt]
\end{tabular}}
\end{center}
\vspace{-0.3cm}
\end{table}

\begin{table}[ht]
\setlength{\tabcolsep}{10pt}
\caption{\small
\textbf{Comparisons between \simper and additional SSL baselines on general periodic learning datasets other than human physiological measurement ones.} When extending to general periodic learning tasks, SSL baselines tailored for human physiological measurement \citep{wang2022self,gideon2021way} no longer provide benefits, and sometimes perform even \emph{worse} than the vanilla supervised learning. In contrast, \simper consistently and substantially exhibits strengths in general periodic learning across all domains.}
\vspace{-1pt}
\label{appendix:table:sota-ssl-algos-other-datasets}
\small
\begin{center}
\adjustbox{max width=0.9\textwidth}{
\begin{tabular}{lcccccc}
\toprule[1.5pt]
 & \multicolumn{2}{c}{RotatingDigits} & \multicolumn{2}{c}{Countix} & \multicolumn{2}{c}{LST} \\
\cmidrule(lr){2-3} \cmidrule(lr){4-5} \cmidrule(lr){6-7}
Metrics & MAE$^\downarrow$ & MAPE$^\downarrow$ & MAE$^\downarrow$ & GM$^\downarrow$ & MAE$^\downarrow$ & $\rho^\uparrow$ \\ \midrule\midrule
\textsc{Supervised} & 0.72 & 28.96 & 1.50 & 0.73 & 1.54 & \textbf{0.96} \\ \midrule
\citep{gideon2021way} & 0.70 & 28.03 & 1.58 & 0.81 & 1.62 & 0.92 \\[1.2pt]
\citep{wang2022self} & 0.77 & 29.44 & 1.68 & 0.94 & 1.64 & 0.89 \\[1.2pt]
\grayrow
\textbf{\textsc{SimPer}} & \textbf{0.20} & \textbf{14.33} & \textbf{1.33} & \textbf{0.59} & \textbf{1.47} & \textbf{0.96} \\
\bottomrule[1.5pt]
\end{tabular}}
\end{center}
\vspace{-0.3cm}
\end{table}

\subsection{Comparisons to SSL Methods in Human Physiological Measurement}
\label{appendix-subsec:sota-ssl-algos-rppg}

In video-based human physiological measurement domain, recent works \citep{wang2022self,gideon2021way} have proposed to leverage contrastive SSL for better learned features and downstream performance in the corresponding application (e.g., heart rate estimation).
They studied specific SSL methods tailored for video-based human physiological measurement, and as a result, many of the proposed techniques therein only apply to that specific domain (e.g., the face detector, the saliency sampler, and the strong assumptions that are derived from the application context, cf. Table 1 in \citep{gideon2021way}). Nevertheless, it is possible to extend the SSL objectives therein to other general periodic learning domains.
In this section, we provide additional experimental results and further discussions, which distinguish \simper from these prior works.

\textbf{Comparisons on the human physiological measurement task.}
We first compare \simper against the aforementioned SSL methods \citep{gideon2021way,wang2022self} on the human physiological measurement task. To provide a fair comparison, we fix all methods to use a simple 3D ConvNet backbone \citep{gulrajani2020domainbed} on SCAMPS, and a TS-CAN backbone \citep{liu2020multi} on UBFC and PURE as stated in Appendix \ref{appendix-sec:exp-settings}.
As Table \ref{appendix:table:sota-ssl-algos-rppg-datasets} demonstrates, \simper outperforms these SSL baselines across all tested human physiology datasets by a notable margin. We break the results out to confirm that they hold regardless of whether we include the customized face saliency module \citep{gideon2021way} or not.

\textbf{Comparisons on other periodic learning tasks.}
We further extend the comparisons to other general periodic learning tasks. We directly apply the SSL objectives in \citep{gideon2021way,wang2022self} to other domains and datasets involving periodic learning, and show the corresponding results in Table \ref{appendix:table:sota-ssl-algos-other-datasets}. The table clearly shows that the SSL objectives in the referenced papers do not provide a benefit in other periodic learning domains, and sometimes perform even \emph{worse} than the vanilla supervised baseline. The above results further emphasize the significance of \simper, which consistently and substantially exhibits strengths in general periodic learning across all domains.

\subsection{Visualization of Learned Features}
\label{appendix-subsec:vis-1d-feats}

\begin{figure}[!t]
\begin{center}
\includegraphics[width=\linewidth]{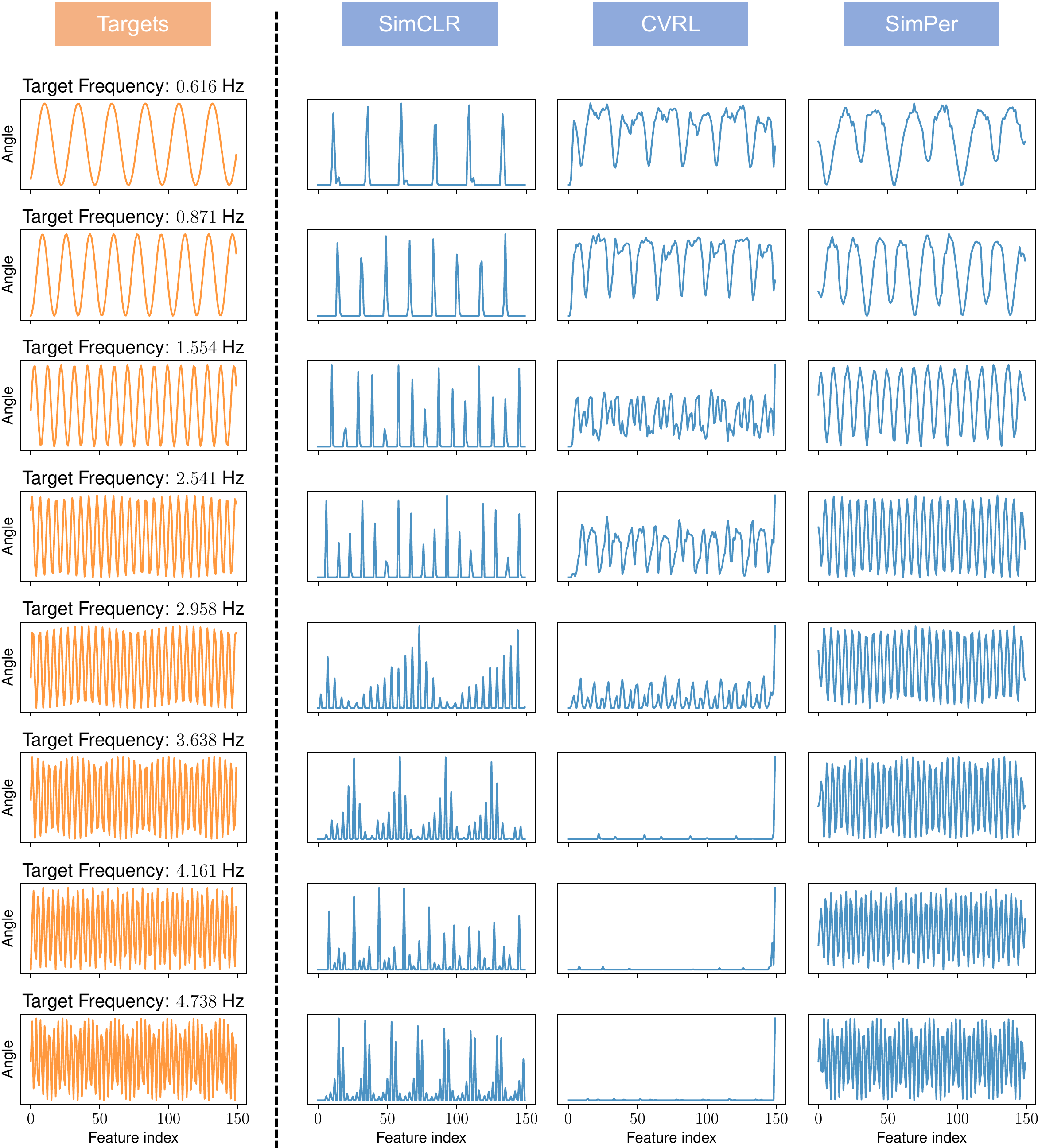}
\end{center}
\vspace{-0.2cm}
\caption{\small \textbf{Visualization of learned periodic representations.} We directly plot the 1-D feature vector of data in the test set of RotatingDigits with different underlying target frequencies (left), via different self-supervised learning methods (right). Existing SSL solutions \citep{qian2021spatiotemporal,chen2020simple} fail to learn meaningful periodic representations, whereas \simper is able to capture the underlying periodicity information.}
\label{appendix:fig:1d-feat-vis}
\end{figure}

Since representations learned in periodic data naturally preserves the periodicity information, we can directly plot the learned 1-D features for visualization.
Fig.~\ref{appendix:fig:1d-feat-vis} shows the learned feature comparison between SimCLR, CVRL, and \simper, together with the underlying periodic information (rotation angle \& frequency) in RotatingDigits.
As the figure verifies, \simper consistently learns the periodic information with different frequency targets, delivering meaningful periodic representations that are robust and interpretable. In contrast, existing SSL methods cannot capture the underlying periodicity, and fail to learn useful representations for periodic learning tasks.

\vspace{0.1cm}

\section{Broader Impacts and Limitations}
\label{appendix-sec:discussions}

\textbf{Limitations.} There are some limitations to our approach in its current form. The \simper features learnt in some cases were not highly effective without certain fine-tuning on a downstream task. This may be explained by the fact that some videos may contain multiple periodic processes (e.g., pulse/PPG, breathing, blinking, etc.). A pure SSL approach will learn features related to all these periodic signals, but not information that is specific to any one.
One practical solution for this limitation could be incorporating the \emph{frequency priors} of the targets of interest.
Precisely, one can filter out unrelated frequencies during \simper pre-training to force the network to learn features that are constrained within a certain frequency range.
We leave this part as future work.

\textbf{Broader Impacts.} While our methods are generic to tasks that involve learning periodic signals, we have selected some specific tasks on which to demonstrate their efficacy more concretely. The measurement of health information from videos has tremendous potential for positive impact, helping to lower the barrier to access to frequent measurement and reduce the discomfort or inconvenience caused by wearable devices. However, there is the potential for negative applications of such technology. Whether by negligence, or bad intention, unobtrusive measurement could be used to measure information covertly and without the consent of a user. Such an application would be unethical and would also violate laws in many parts of the world\footnote{\url{https://www.ilga.gov/legislation/ilcs/ilcs3.asp?ActID=3004&ChapterID=57}}.
It is important that the same stringent measures applied to traditional medical sensing are also applied to video-based methods.
We will be releasing code for our approach under a Responsible AI License (RAIL)~\citep{contractor2022behavioral} to help practically mitigate unintended negative behavioral uses of the technology while still making the code available.




\end{document}